\documentclass{article}

\usepackage{microtype}
\usepackage{graphicx}
\usepackage{subcaption}
\usepackage{booktabs} 
\usepackage{hyperref}
\usepackage{algpseudocode}

\usepackage[preprint]{icml2026}

\usepackage{amsmath}
\usepackage{amssymb}
\usepackage{mathtools}
\usepackage{amsthm}
\usepackage{xcolor}
\usepackage{pifont}
\usepackage{bm}
\usepackage[hang,flushmargin]{footmisc}
\usepackage{color, colortbl}

\usepackage[capitalize,noabbrev]{cleveref}

\theoremstyle{plain}

\theoremstyle{definition}

\theoremstyle{remark}

\usepackage[textsize=tiny]{todonotes}

\icmltitlerunning{FastPhysGS: Accelerating Physics-based Dynamic 3DGS Simulation via 
Interior Completion and Adaptive Optimization}

\begin{document}

\twocolumn[
  \icmltitle{FastPhysGS: Accelerating Physics-based Dynamic 3DGS Simulation via \\
Interior Completion and Adaptive Optimization}



  
  \icmlsetsymbol{equal}{*}

  \begin{icmlauthorlist}
    \icmlauthor{Yikun Ma}{sch1}
    \icmlauthor{Yiqing Li}{sch1}
    \icmlauthor{Jingwen Ye}{comp}
    \icmlauthor{Zhongkai Wu}{comp}
    \icmlauthor{Weidong Zhang}{comp}
    \icmlauthor{Lin Gao}{zky1,zky2}
    \icmlauthor{Zhi Jin}{sch1,lab1,lab2}
  
  \end{icmlauthorlist}

  \icmlaffiliation{sch1}{School of Intelligent Systems Engineering, Shenzhen Campus of Sun Yat-sen University, Shenzhen, Guangdong 518107, P.R.China.} 
  \icmlaffiliation{comp}{Anonymous Institution.}
  \icmlaffiliation{zky1}{Institute of Computing Technology, Chinese
    Academy of Sciences, Beijing 100190, China.}
  \icmlaffiliation{zky2}{University of
    Chinese Academy of Sciences, Beijing 101408, China.}
  \icmlaffiliation{lab1}{Guangdong Provincial Key Laboratory of Fire Science and Intelligent Emergency Technology, Shenzhen 518107, P.R.China.}
  \icmlaffiliation{lab2}{Guangdong Provincial Key Laboratory of Robotics and Digital Intelligent Manufacturing Technology, Guangzhou 510535, PR China.}
  \icmlcorrespondingauthor{Zhi Jin}{jinzh26@mail.sysu.edu.cn}

  \icmlkeywords{Machine Learning, ICML}

  \vskip 0.3in
]



\printAffiliationsAndNotice{}  

\begin{abstract}
    Extending 3D Gaussian Splatting (3DGS) to 4D physical simulation remains challenging. 
    Based on the Material Point Method (MPM), existing methods either rely on manual parameter tuning or distill dynamics from video diffusion models, 
    limiting the generalization and optimization efficiency. 
    Recent attempts using 
    LLMs/VLMs suffer from a text/image-to-3D perceptual gap, yielding unstable physics behavior. 
    In addition, they often ignore the surface structure of 3DGS, leading to implausible motion. 
    We propose FastPhysGS, a fast and robust framework for physics-based dynamic 3DGS simulation: 
    (1) Instance-aware Particle Filling (IPF) with Monte Carlo Importance Sampling (MCIS) to efficiently populate interior particles while preserving geometric fidelity; 
    (2) Bidirectional Graph Decoupling Optimization (BGDO), an adaptive strategy that rapidly optimizes material parameters predicted from a VLM. 
    Experiments show FastPhysGS achieves high-fidelity physical simulation in 1 minute using only 7 GB runtime memory, 
    outperforming prior works with broad potential applications. 
\end{abstract}

\begin{figure}[!t]
    \centering
    \includegraphics[width=0.92\linewidth]{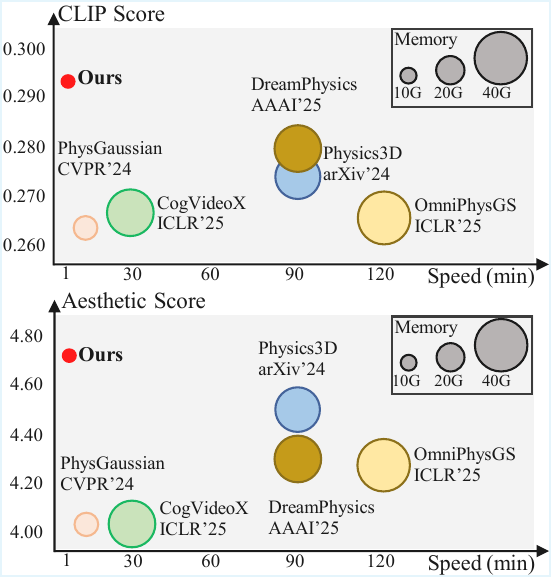}
    \vspace{-0.22cm}
    \caption{We propose FastPhysGS, an efficient and robust physics-based dynamic 3DGS simulation framework. 
    Input a 3DGS scene, our method requires only 7 GB of memory and completes complex dynamic simulations within 1 minute, making it practical to generate real-time 4D physics-aware dynamics.}
    \label{fig:first}
    \vspace{-0.3cm}
\end{figure}

\section{Introduction}
\label{sec:intro}

Recently, remarkable progress in 3D reconstruction \cite{mildenhall2020nerf, kerbl20233d, wang2025vggt} and generation \cite{pooledreamfusion, wang2023prolificdreamer, hollein2023text2room, ma2024fastscene} has significantly advanced applications in gaming, virtual reality, and robotics. 
However, generating realistic, dynamically consistent, and spatio-temporally coherent 4D contents remain challenging. 
With the progress in video generation models 
\cite{brooks2024video, yangcogvideox, kong2024hunyuanvideo}, 
several studies have explored constructing 4D content that reflects real-world dynamics \cite{ren2023dreamgaussian4d, zeng2024stag4d, ren2024l4gm}. 
These methods typically leverage dynamic priors derived from video diffusion models to predict 4D deformations and subsequently reconstruct spatio-temporal scenes. 
Nevertheless, they inherently lack explicit physical constraints, often resulting in motions that violate physical laws and exhibit spatio-temporal inconsistencies of 4D content.

To address these issues, PhysGaussian \cite{xie2024physgaussian} first integrates the Material Point Method (MPM) \cite{stomakhin2013material} with 3D Gaussian Splatting (3DGS) \cite{kerbl20233d}, leveraging an explicit Lagrangian-Eulerian grid representation to simulate physically plausible dynamics in 3DGS. 
However, it requires manually specifying the physical material properties for each scene, which demands substantial expert knowledge. 
To mitigate this, subsequent methods \cite{liu2024physics3d, linomniphysgs, huang2025dreamphysics} adopt Score Distillation Sampling (SDS) \cite{pooledreamfusion}  to distill  material properties from video diffusion models. 
However, this way is computationally expensive and relies on the motion priors from video diffusion models. 
In contrast, recent methods \cite{lin2024phys4dgen, zhao2024efficient, chen2025physgen3d} employ LLMs/VLMs to predict physical parameters due to their reasoning capabilities. 
Nevertheless, a perceptual gap exists between text/images and native 3D content, which leads to unstable simulation results. 
Furthermore, existing methods overlook the hollow characteristic of 3DGS, 
resulting in unrealistic physical behaviors. 

In summary, existing methods face three key challenges: 
\textbf{(1)} 
Neglecting the hollow 3DGS structure, 
while lacking the ability to distinguish instance-level filling. 
\textbf{(2)} 
Relying on manually designed parameters or diffusion distillation, leading to inadequate generalization and inefficient simulation. 
\textbf{(3)} 
Employing large models to predict physical parameters, leading to a perception gap with the native 3DGS content.

\begin{figure}[!t]
    \centering
    \includegraphics[width=0.92\linewidth]{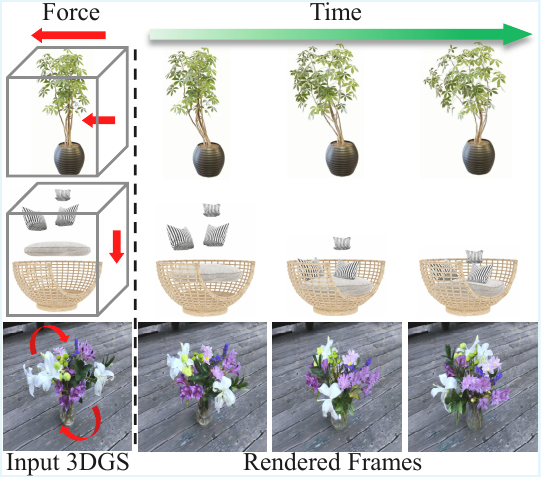}
    \vspace{-0.1cm}
    \caption{FastPhysGS supports various physical behaviors including movement, collision, tearing, rotation, swaying  
    across diverse materials, such as sand, rubber, jelly, water and elastomers.}
    \label{fig:first_fig}
    \vspace{-0.3cm}
\end{figure}

To address the aforementioned challenges, we propose FastPhysGS with two stages: 
\textbf{(1)} 
Considering the spatial structure of 3DGS, 
we propose Instance-aware Particle Filling (IPF). 
Our approach begins with segmenting the 3DGS objects, and leverages ray-casting guided by an occupancy field to estimate initial internal particles. 
To handle complex geometries filling, 
such as irregular surfaces and concave regions, 
we then design the Monte Carlo Importance Sampling (MCIS). 
\textbf{(2)} 
For adaptive and efficient refinement of physical parameters predicted by a VLM, 
we propose Bidirectional Graph Decoupling Optimization (BGDO). 
By exploring the stress-strain mechanics, 
BGDO leverages stress gradients and deformation to efficiently optimize parameters. 
Extensive experiments demonstrate that compared to other physical dynamic simulation methods, FastPhysGS generates plausible motions rapidly and robustly. 
As shown in Figures \ref{fig:first} and \ref{fig:first_fig}, 
our method achieves state-of-the-art results across a wide range of physical simulations and material types, only requires \textcolor{red}{7 GB} running memory in \textcolor{blue}{1 minute}, which demonstrates outstanding practicality and broad potential application. 
Our main contributions are as follows: 

\begin{itemize}
    \item 
    \textbf{Real-Time, Memory-Efficient  Dynamic 3DGS Simulation:} 
    FastPhysGS establishes a novel paradigm for physics-based dynamic 3DGS, achieving high-fidelity, multi-materials simulation with  
    remarkable efficiency and physical plausibility. 
    \item 
    \textbf{Instance-aware Complete Geometry Representation:}
    We propose IPF with MCIS, 
    which explicitly fills the hollow 3DGS, providing a geometrically complete and stable foundation for physical simulation. 
    
    \item 
    \textbf{Perception-to-Physics optimization:}
    We present BGDO, a novel adaptive refinement mechanism  to rapidly correct VLM-predicted material parameters. 
    
\end{itemize}

The rest of this paper is organized as follows: 
Sec.~\ref{sec:related} reviews the related work. Sec.~\ref{sec:Preliminaries} introduces the preliminaries of 3DGS and MPM.
Sec.~\ref{sec:method} presents the details of our proposed method.
Sec.~\ref{sec:experment} provides experimental analysis and results.
Finally, Sec.~\ref{sec:conclusion} concludes the paper. 

\section{Related Work}
\label{sec:related}

\subsection{4D Content Generation}
\label{4dgen}
4D content creation has grown significantly across various applications.  
With the success of video generation models  \cite{brooks2024video, yangcogvideox, kong2024hunyuanvideo}, 
several methods propose to generate 4D content using video priors. 
MAV3D \cite{singer2023text} first employs temporal Score Distillation Sampling (SDS) \cite{pooledreamfusion} from the text-to-video diffusion model. 
Animate124 \cite{zhao2023animate124} proposes an image-to-4D coarse-to-fine framework, 
while DreamGaussian4D \cite{ren2023dreamgaussian4d} utilizes pre-generated videos to supervise the deformation of 3DGS. 
However, the diffusion dynamics may be physically inaccurate, 
and directly applying them leads to inplausible results.

\subsection{Physics-Based Dynamic Simulation}
To address the issues mentioned in Sec.\ref{4dgen}, 
recent works \cite{xie2024physgaussian, linomniphysgs, zhang2024physdreamer, huang2025dreamphysics, liu2024physics3d} introduce physics-based simulation. 
PhysGaussian \cite{xie2024physgaussian} first coupled 3DGS \cite{kerbl20233d}
with continuum mechanics simulation via MPM \cite{stomakhin2013material}. 
However, it requires manual configuration of physical parameters for each scene, 
causing inconvenient simulation. 
Subsequent approaches focus on automating MPM material perception. 
For instance, DreamPhysics \cite{huang2025dreamphysics} and Physics3D \cite{liu2024physics3d} 
leverage video generation models to estimate physical material parameters by SDS. 
However, it is time-consuming and relies on generative priors. 
To accelerate simulation, methods like PhysSplat \cite{zhao2024efficient} and Phys4DGen \cite{lin2024phys4dgen} utilize LLMs/VLMs to predict parameters. 
However, they rely on the output of large models, creating a perception gap with the actual 3D environment, leading to instable simulation. 
Moreover, existing works often overlook the hollow interior of 3DGS, which distorts the computation of particle stresses. 
Although PhysGaussian \cite{xie2024physgaussian} introduces particle filling, 
its global processing makes it difficult to distinguish object instances and different materials. 

\section{Preliminaries}
\label{sec:Preliminaries}
\subsection{3D Gaussian Splatting}
3DGS \cite{kerbl20233d} represents a scene as a collection of 3D gaussians, each defined by a center $\bm{\mu}$ and a covariance matrix $\bm{\Sigma}$. 
The gaussian function at position $\bm{x}$ is given by:
\begin{equation}
G(\bm{x}) = \exp\left(-\frac{1}{2} (\bm{x} - \bm{\mu})^\top \bm{\Sigma^{-1}} (\bm{x} - \bm{\mu})\right), 
\end{equation}
where the covariance matrix $\bm{\Sigma}$ is decomposed as $\bm{\Sigma = R S S^\top R^\top}$, where $\bm{S}$ is a diagonal scaling matrix and $\bm{R}$ is a rotation matrix. 
During rendering, view transformation $\bm{W}$ is applied, and the 2D projected covariance is computed using the Jacobian $\bm{J}$:
$ \bm{\Sigma'} = \bm{J W \Sigma W^\top J^\top.}$ 
The pixel color $\bm{Col}$ is computed by blending $N$ overlapping gaussians:
\begin{equation}
    \bm{Col} = \sum_{i=1}^{N} col_i \alpha_i \prod_{j<i} (1 - \alpha_j),
\end{equation}
where $col_i$ and $\alpha_i$ denote the color and opacity of the $i$-th gaussian, respectively. 

\subsection{Material Point Method}
MPM \cite{stomakhin2013material} is a widely used Lagrangian-Eulerian framework for physics-based simulation of continuum materials. 
In MPM, each time step consists of three sequential stages: 
(1) \textit{Particle-to-Grid (P2G)}, where particle mass $m$ and momentum are mapped to a background grid; 
(2) \textit{Grid update}, where the discretized momentum equations are solved under applied internal and external forces $\textbf{f}$; 
(3) \textit{Grid-to-Particle (G2P)}, where updated velocities, positions, and deformation gradients are interpolated back to the particles. 
We adopt MLS-MPM \cite{hu2018moving}, a MPM variant that improves momentum conservation  via local affine velocity fields, 
and follow PhysGaussian to define each gaussian kernel as the time-dependent state: 
\begin{equation}
    \bm{x}_i(t) = \Delta(\bm{x}_i, t), \quad \bm{\Sigma_i(t)} = \bm{F}_i(t) \bm{\Sigma_i} \bm{F}_i(t)^\top,
\end{equation}
where $\Delta(\cdot, t)$ and $\bm{F}_i(t)$ denote coordinate deformation and deformation gradient at timestep $t$. 

\begin{figure}[!b]
    \centering
    \includegraphics[width=0.98\linewidth]{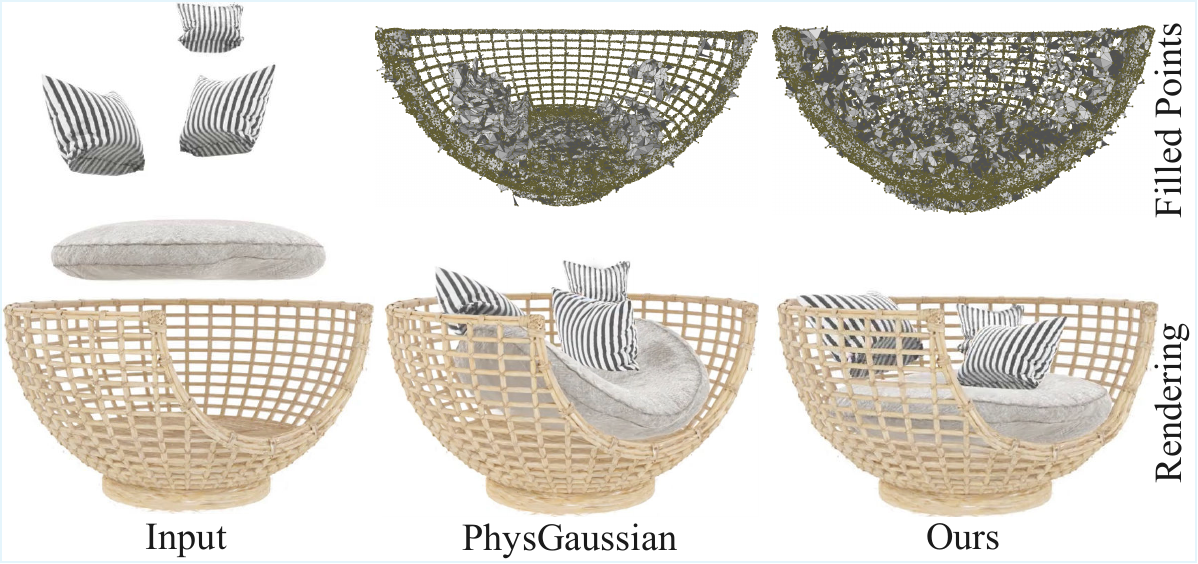}
    \vspace{-0.2cm}
    \caption{We extract the filled points as meshes for better visualization. 
    PhysGaussian incorrectly fills the hollow region of the wicker basket, causing the mat to warp upward, while our method achieves accurate instance-aware filling.} 
    \label{fig:fill_com}
\end{figure}

\section{Method}
\label{sec:method}

\begin{figure*}[htbp]
    \centering
    \includegraphics[width=0.97\linewidth]{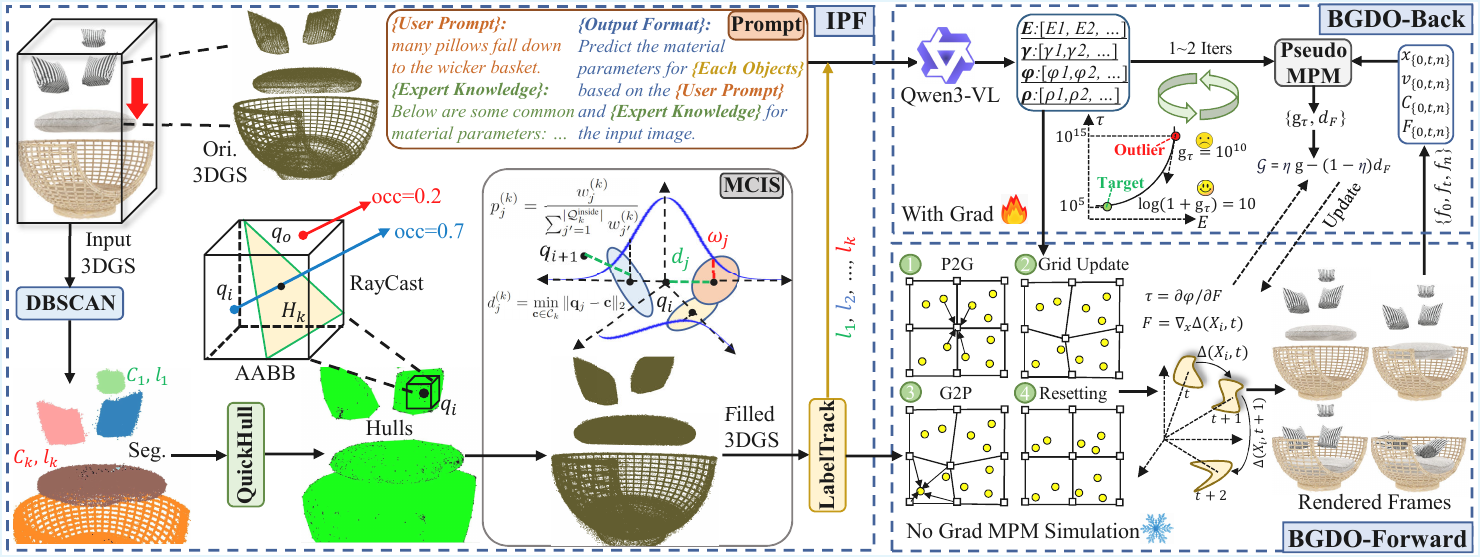}
    \vspace{-0.2cm}
    \caption{The pipeline of our method. 
    The first stage IPF rapidly fills the interior 3DGS particles, 
    while MCIS is designed to identify crucial points and  handle complex geometries. 
    The second stage BGDO is proposed to adaptively optimize MPM parameters. 
    Overall, our method generates complete and realistic 4D physical dynamics in 1 minute, showcasing great potential in practical applications. 
    }
    \vspace{-0.3cm}
    \label{fig:pipeline}
\end{figure*}

The overall pipeline of FastPhysGS is illustrated in Figure~\ref{fig:pipeline}, consisting of two stages: IPF and BGDO. 
Specifically, IPF populates the interior 3DGS particles, leveraging MCIS to handle complex geometry filling. 
Subsequently, BGDO employs the bidirectional graph decoupling strategy, which performs forward MPM simulation and backward optimization to refine physical parameters predicted from the VLM. 
Details of IPF and BGDO are described in Sec. \ref{sec_IPF} and \ref{sec_BGDO}. 

\subsection{Instance-aware Particle Filling}\label{sec_IPF}
3DGS essentially reconstructs the surface appearance with empty internal structure. 
Therefore, it is prone to collapse during simulation, as the internal stress contribution is minimal. 
PhysGaussian \cite{xie2024physgaussian} first utilizes the opacity field to determine internal points by checking whether a ray passes through grids with different opacities. 
However, this global filling makes it difficult to assign distinct materials to multiple objects (e.g., granting each object  independent physical properties), 
while lacking reasonable adaptation for complex structures. 
For example, Figure \ref{fig:fill_com} shows that PhysGaussian incorrectly fills the hollow region of the bamboo basket, causing the mat sides to be erroneously lifted upward. 

To tackle these problems, we propose IPF to efficiently populate 3DGS particles with instance-aware capability. 
As shown in Figure \ref{fig:pipeline}, 
let the original 3DGS positions $\bm{\mathcal{P}}$ be: 
\begin{equation}
    \bm{\mathcal{P}} = \{ \bm{p}_i \in \mathbb{R}^3 \}_{i=1}^N , 
\end{equation}
where $\bm{p}_i$ denotes the mean of each gaussian, 
and $N$ represents the total number of gaussians. 
Specifically, we apply DBSCAN clustering \cite{ester1996density} to obtain 
$K$ object instances for subsequent instance-aware MPM simulation: 
\begin{align}
     & \{ \bm{\mathcal{C}_k},  l_k \ \}_{k=1}^K = \text{DBSCAN}(\bm{\mathcal{P}}, r),
    \\
     & \bm{\mathcal{C}_k}  = \{ \bm{c}_j^{(k)} \in \mathbb{R}^3 \}, \quad k = 1, 2, \dots, K ,
\end{align}
where $\bm{\mathcal{C}_k}$ denotes the $k$-th point cluster, 
$l_k$ indicates the label of each cluster,
$r$ denotes the cluster radius,
$\bm{c}_j$ represents the $j$-th point within the cluster $\bm{\mathcal{C}_k}$. 

To rapidly obtain the interior gaussian points of each cluster $\bm{\mathcal{C}_k}$, we independently compute the convex hull utilizing Quickhull \cite{barber1996quickhull}:
$\bm{\mathcal{H}_k} = \text{Quickhull}(\bm{\mathcal{C}_k}), $
where $\bm{\mathcal{H}_k}$ are the triangular meshes enclosing the points in $\bm{\mathcal{C}_k}$. 
For all possible candidate filling points $\bm{\mathcal{Q}_k}$ of each $\bm{\mathcal{C}_k}$, we construct AABB bounding  boxes $\bm{\mathcal{B}_k}$ from $\bm{\mathcal{H}_k}$:
\begin{align}
    & \bm{\mathcal{Q}_k} = \left\{ \bm{q}_j \sim \text{Uni}(\bm{\mathcal{B}_k}) \right\}_{j=1}^{M}, \quad M \gg |\mathcal{C}_k|, 
    \\ 
    & \bm{\mathcal{B}_k} = [\bm{b}_{\min}^{(k)}, \bm{b}_{\max}^{(k)}] \subset \mathbb{R}^3 , 
\end{align}
where $Uni$ represents uniform random sampling within the boxes, 
and $\bm{b}$ denotes the boundary value of the hull $\bm{\mathcal{H}_k}$.

To efficiently determine whether a candidate point $\bm{q}$ lies inside the convex hull $\bm{\mathcal{H}_k}$, we compute a 3D occupancy field $\text{occ}(\bm{q}; \bm{\mathcal{H}_k}) \in [0, 1]$, which employs fast ray-casting \cite{1990Display} to coarsely select the interior points based on occupancy probability threshold (experimentally set to 0.6): 
\begin{equation}
    \bm{\mathcal{Q}_k^{\text{inside}}} = \left\{ \bm{q}_j \in \bm{\mathcal{Q}_k} \mid \text{occ}(\bm{q}_j; \bm{\mathcal{H}_k}) >0.6 \right\} .
\end{equation} 

However, relying solely on the occupancy field results in incorrect filling for some irregularly shaped objects (e.g. Figure \ref{fig:MCIS} illustrates the bamboo basket below, 
whose center is hollow and should not be filled). 
To address this, we propose the Monte Carlo Importance Sampling (MCIS) \cite{kocsis2006bandit}  strategy:
\begin{equation}
\begin{split}
&\mathbb{E}_p[f] = \int f(\bm{x})\, p(\bm{x})\, d\bm{x} = \int f(\bm{x})\, \frac{p(\bm{x})}{q(\bm{x})}\, q(\bm{x})\, d\bm{x} \\
&\approx \frac{1}{n} \sum_{j=1}^{n} f(\bm{x}_j) \cdot \frac{p(\bm{x}_j)}{q(\bm{x}_j)} = \frac{1}{n} \sum_{j=1}^{n} f(\bm{x}_j) \cdot w_j \ , 
\end{split}
\end{equation}
this is a Monte Carlo expected value of any function $f(\bm{x})$ under 
distribution $p(\bm{x})$. 
The key insight lies in designing a non-uniform probability density function $p(x)$ to assign higher probabilities to important points from $\bm{\mathcal{Q}_k}$. 
We sample $\bm{x}_j \sim q(\bm{x})$, and provide the detailed importance sampling steps of MCIS: 
\begin{figure}[t]
    \centering
    \includegraphics[width=0.81\linewidth]{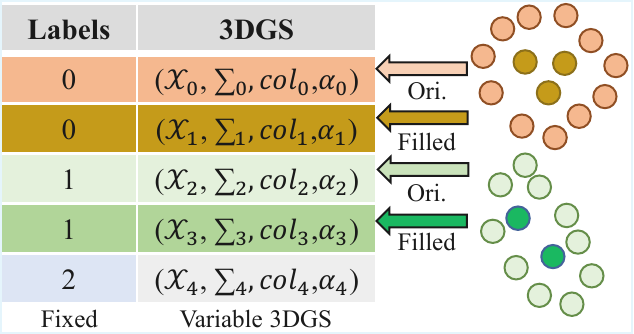}
    \vspace{-0.15cm}
    \caption{We use a contiguous memory block to store labels for tracking the dynamically varying 3DGS properties. 
    }
    \vspace{-0.15cm}
    \label{fig:label_track}
\end{figure}

\textbf{1) Proximity-aware distance metric.} 
For each interior candidate point $\bm{q}_j \in \bm{\mathcal{Q}_k^{\text{inside}}}$ , we compute its minimal Euclidean distance to the observed surface proxy $\bm{\mathcal{C}_k}$  (e.g., a set of anchor gaussians or boundary samples):
\begin{equation}
d_j^{(k)} = \min_{\bm{c} \in \bm{\mathcal{C}_k}} \|\bm{q}_j - \bm{c}\|_2.
\end{equation}

\textbf{2) Gaussian-derived importance weighting.}  
Considering the distribution characteristic of 3DGS, we map distances into importance weights via an gaussian kernel, which exponentially attenuates influence with increasing distance: 
\begin{equation}
w_j^{(k)} = \max\left(\exp\left( -\frac{(d_j^{(k)})^2}{2\sigma^2} \right),\ \epsilon\right), \ \epsilon = 10^{-6}, 
\end{equation}
where $\sigma$ controls the locality of influence (empirically set to 0.02, see Sec. \ref{sec_aba} for more details).

\textbf{3) Normalized importance distribution.}  
The unnormalized weights $w_j^{(k)}$ are converted into a discrete probability function over $ \bm{\mathcal{Q}_k^{\text{inside}}}$ :
\begin{equation}
p_j^{(k)} = \frac{w_j^{(k)}}{\sum_{j'=1}^{|\bm{\mathcal{Q}_k^{\text{inside}}}|} w_{j'}^{(k)}}.
\end{equation}

\textbf{4) Stochastic sampling.}  
We obtain $n_k$ points via multinomial sampling according to $p_j^{(k)}$, yielding a refined subset:
\begin{equation}
\bm{\mathcal{S}_k} = \{ \bm{s}_i^{(k)} \}_{i=1}^{n_k}, \quad \text{with} \ \mathbb{P}(\bm{s}_i^{(k)} = \bm{q}_j) = p_j^{(k)}, 
\end{equation}
where $\mathbb{P}$ represents the sampling probability,  
indicating that each $\bm{q}_j$ is sampled by an importance distribution. 

\textbf{5) Aggregation of filled points.}  
The final outputs are constructed by uniting the original $\bm{\mathcal{P}}$ with the sampled points: 
\begin{equation}
\bm{\mathcal{P}_{\text{filled}}} = \bigcup_{k=1}^{K} \bm{\mathcal{S}_k}, \quad \bm{\mathcal{P}_{\text{all}}} = \bm{\mathcal{P}} \cup \bm{\mathcal{P}_{\text{filled}}}.
\end{equation}

\textbf{6) Appearance suppression for structural fidelity.} 
To prevent visual artifacts from $\bm{\mathcal{P}_{\text{filled}}}$, 
we nullify their appearance contribution by setting opacity $\alpha = 0$, retaining only positional occupancy. 

Finally, to consistently retrieve dynamic gaussian parameters from a fixed label memory $l_k$, 
we propose label tracking 
as illustrated in Figure \ref{fig:label_track}. 
In conclusion, IPF with MCIS fills the interior 3DGS particles, providing a geometrically complete structure foundation for subsequent MPM simulation. 

\subsection{Bidirectional Graph Decoupling Optimization}
\label{sec_BGDO}
\textbf{Optimization Target Analysis.}
To conveniently obtain the physical parameters for MPM simulation, 
we employ Qwen3-VL~\cite{Qwen3-VL} to  obtain an initial prediction $\{\rho, \nu, \Phi, E\}$ for each object instance $l_k$. 
Density $\rho$ mainly influences the inertial term and has a relatively smooth effect. 
Poisson’s ratio $\nu$ lies in a narrow range (0 $\sim$ 0.5) and yields stable predictions thanks to rich pre-defined expert priors provided to the VLM. 
The energy density model $\Phi$ adopts a pre-defined form (e.g., Fixed Corotated Elasticity \cite{stomakhin2012energetically}, StVK Elasticity \cite{barbivc2005real}, Drucker-Prager Plasticity \cite{drucker1952soil}, and Fluid Plasticity \cite{stomakhin2014augmented}), 
and we classify it by VLM. 
Thus, we optimize Young’s modulus 
$E$, as it dominantly governs material stiffness.  
However, a perceptual gap between model reasoning and the native 3D space leads to unsuitable simulation, necessitating an optimization method to address this issue.

Specifically, we first derive the computational graph from $E$ to the rendered frames $\bm{I}$: 
\begin{equation}
\bm{I} \!\leftarrow\! \{\bm{\mathcal{X}}, \bm{\Sigma}, \alpha, col\} \!\leftarrow\! \{\bm{x}, \bm{v}, \bm{C}, \bm{F}\} \!\leftarrow\! \boldsymbol{\tau} \!\leftarrow\! \Phi \!\leftarrow\! E,
\end{equation}
where $\{\bm{\mathcal{X}}, \bm{\Sigma}, \alpha, col\}$ are the 3DGS parameters, and 
$\{\bm{x}, \bm{v}, \bm{C}, \bm{F}\}$ represent the MPM particle states (position, velocity, affine momentum tensor, and deformation gradient),
$\boldsymbol{\tau}$ is the first Piola--Kirchhoff stress. 
Since 3DGS parameters are directly generated from the MPM simulation, 
the rendered frames are merely visual representation of the underlying physical states. 
Therefore, we further analyze the numerical computations in the momentum exchange stage of MPM:  
\begin{equation}
\begin{aligned}[b]
    & m_i^t \bm{v}_i^t ={} \sum_p N(\bm{x}_i - \bm{x}_p^t) \Bigl[ m_p \bm{v}_p^t + \\
    &  \Bigl( m_p \bm{C}_p^t - \frac{4}{(\Delta x)^2} \Delta t V_p \frac{\partial \Phi}{\partial \bm{F}} \bm{F}_p^{tT} \Bigr) (\bm{x}_i - \bm{x}_p^t) \Bigr] + \bm{f}_i^t,
\end{aligned}
\end{equation}
where $\partial \Phi / \partial \bm{F}$ denotes the particles stress $\bm{\tau}$, which captures the material-level forces in MPM.  
To conveniently control different material properties for multi-objects, 
we set the $\bm{\tau}$ as a variable term changing with different $E$ in K lables: 
\begin{equation}\label{PK_eq_sum}
\bm{\tau}(\bm{F}, E) = \sum_{k=1}^{K} \frac{\partial \Phi_k (\bm{F}_k, E_k)}{\partial \bm{F}_k}.
\end{equation}
Based on the above analysis, 
we aim to efficiently optimize $E$ by leveraging $\{\bm{\tau}, 
\bm{F}\}$ as the physical plausibility signal, with two primary objectives: 
\textbf{(1) Time consumption:} avoid optimization during MPM simulation process, as current frame is determined by the previous frame, the full computation is highly time-consuming; 
\textbf{(2) Memory accumulation:} exclude simulation process from the computation graph, as it accumulates gradients across frames, 
leading to excessive GPU memory usage and out-of-memory errors. 

\textbf{Optimization Algorithm.}
Therefore, we propose BGDO to separate the forward simulation from backward optimization. 
As shown in Figure \ref{fig:pipeline}, 
we first perform a gradient-free forward MPM with a temporal and causal process, 
where subsequent frames are only influenced by prior states. 
Thus, we record only three key frames (initial frame~$\bm{f_0}$, an intermediate frame~$\bm{f_t}$, and the final frame~$\bm{f_n}$) with attributes $\{\bm{x}, \bm{v}, \bm{C}, \bm{F}\}$. 
This provides effective priors for optimization while minimizing storage costs.

Then, we enable gradient computation in backward graph, and perform a \textbf{pseudo} simulation process, which preserves only the gradient pathways without actual simulation. 
To optimize $E$ using $\bm{\tau}$ and $\bm{F}$, we leverage the stress gradient, while supervising material deformation by measuring the Frobenius norm \cite{van1996matrix} as guidance: 
\begin{equation}
    g_{\tau} = \frac{1}{3} \sum_{i \in \{0, t, n\}} \frac{\partial \|\bm{\tau_i}\|}{\partial  E_i}, \ d_{\bm{F}} = \frac{1}{3} \sum_{i \in \{0, t, n\}} \left\| \bm{F_i} - \mathbf{I} \right\|_F. 
\end{equation}
However, the stress gradient exhibits an wide dynamic range ($\sim 10^{8}$ to $10^{20}$), 
causing numerical instability if used directly.
Thus, we compress its magnitude via a logarithmic transform $g = \log(1 + g_\tau)$, and the optimization equation $\mathcal{G}$ is formulated as a weighted combination of $g$ and $d_{\bm{F}}$: 
\begin{equation}
    \mathcal{G} = \eta \cdot g - (1-\eta) \cdot d_{\text{F}}, 
    \ \eta=min(1, 0.1 \cdot log(1+E)), \label{eq_target}
\end{equation} 
where $\eta$ is the guidance weight, which is proportional to $E$. 
Intuitively, an overly large $E$ amplifies abnormal stress gradients, increasing the influence of the first term $\eta \cdot g$, while an excessively small $E$ results in softer material and strengthens the effect of the deformation gradient in the second term $(1-\eta) \cdot d_{\text{F}}$. 
The balance between these two signals enables stable and adaptive material optimization.
Finally, the update of Young's modulus $ E $ is defined as: 
\begin{equation}
log (E) \xleftarrow{} log (E) - \mathcal{G}  \ .
\end{equation}
Overall, BGDO is formulated from the perspective of numerical computation within the simulation, 
allowing for adaptive and rapid rectification of undesirable predicted parameters in only 1–2 iterations. 
IPF takes about \textbf{22s}, BGDO forward simulation takes \textbf{39s}, and the backward optimization occurs almost instantaneously, as shown in Table \ref{tab:timememory}.

\begin{table}[htbp]
\centering
\small
\setlength{\tabcolsep}{4pt}
\caption{
The running time and memory across different stages. 
The results are averaged over our entire experiment dataset. 
}
\vspace{-0.2cm}
\label{tab:timememory}
\begin{tabular}{c|c|c|c|c}
\hline
Method      & IPF   & BGDO-Forward  &  BGDO-Back & Full \\
\hline
Time        & 22 s  & 39 s   & $<$1 s & 1 min  \\
\hline
Memory      & 1.6 GB  & 7 GB    & 7 GB & 7 GB  \\
\hline
\end{tabular}
\vspace{-0.4cm}
\end{table}

\begin{table*}[!t]
\centering
\caption{
Quantitative comparisons. 
The CLIP Score (CS) measures the semantic alignment between text and frames, 
while the Aesthetic Score (AS) evaluates visual quality. 
Semantic Adherence (SA) measures the video-text alignment fidelity, 
while Physical Commensense (PC) measures 
evaluates the compliance with real-world physics. 
The best results are highlighted in \textbf{bold}. 
}
\vspace{-0.2cm}
\label{tab:com}
\begin{tabular}{c|c|c|c|c|c|c|c|c}
\hline
Method       & Pub. & Auto Param. & CS$\uparrow$  & AS$\uparrow$  & SA$\uparrow$ & PC$\uparrow$ &  Memory (GB)$\downarrow$  &  Time (min)$\downarrow$ \\
\hline
CogVideoX1.5-5B & ICLR'25 & -               & 0.269  & 4.04  & 0.44  & 0.21  &  33      & 24  \\
\hline
Veo-3.1           & Google'25  & -          &  0.291  & \textbf{4.90}  & 0.81  & 0.49  & - & -\\
\hline
PhysGaussian    & CVPR'24 & \ding{55}         & 0.267  & 4.06 &  0.69 & 0.55  &  12       & 2 \\
\hline
Physics3D       & arXiv'24 & \ding{52}        & 0.275  & 4.48  &  0.81  &  0.53  &  30       &  90  \\
\hline
DreamPhysics    & AAAI'25 & \ding{52}        & 0.279  & 4.38   &  0.75 &  0.51 & 35        & 90  \\
\hline
OmniPhysGS      & ICLR'25 & \ding{52}        & 0.265  & 4.35  & 0.63  & 0.45  &  40       &  120  \\
\hline
\rowcolor{red!10}
Ours            & - & \ding{52}        & \textbf{0.292}  & 4.71  & \textbf{0.87}  & \textbf{0.61}  & \textbf{7}         & \textbf{1}  \\
\hline
\end{tabular}
\vspace{-0.3cm}
\end{table*}

\begin{figure*}[!t]
    \centering
    \includegraphics[width=0.95\linewidth]{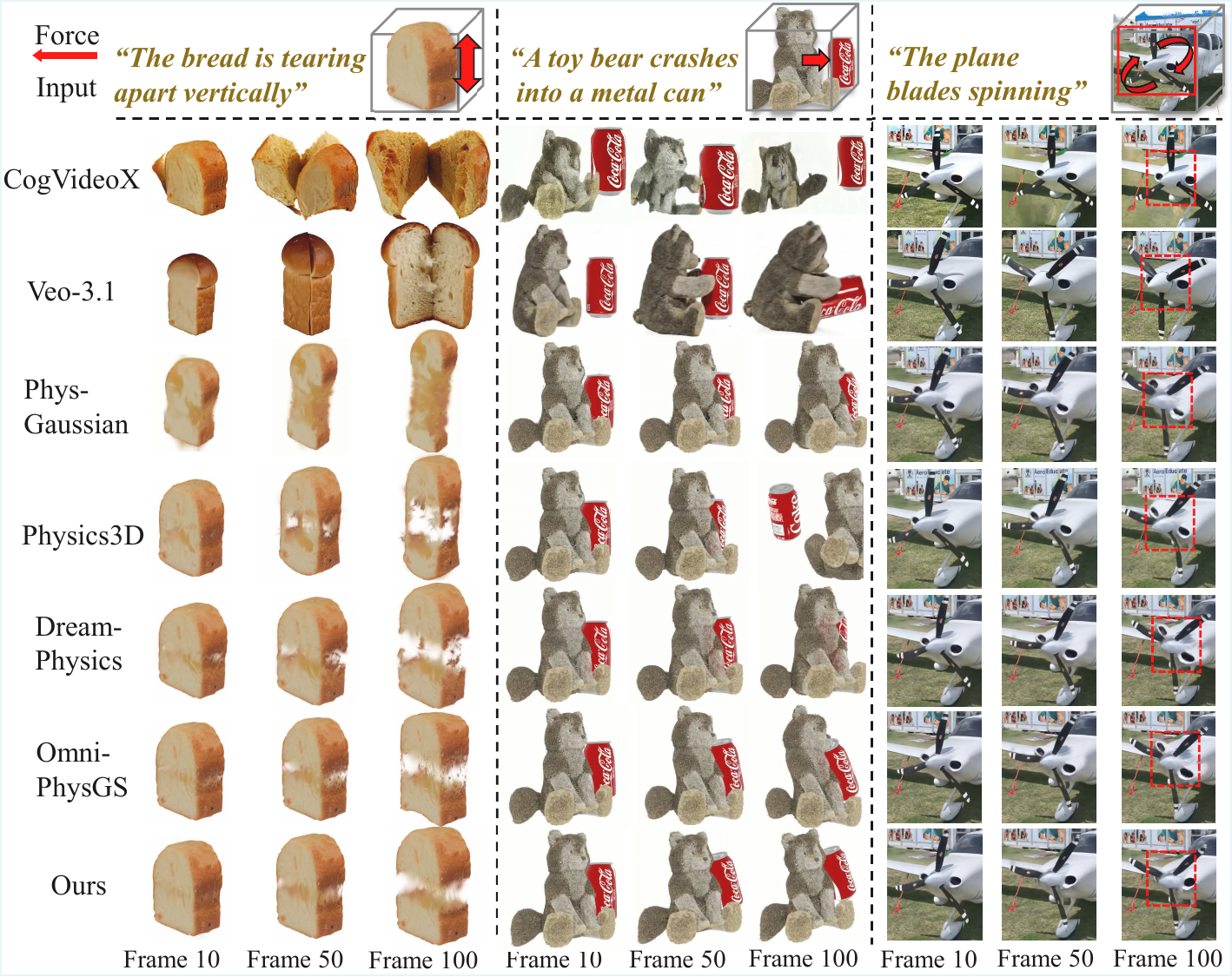}
    \vspace{-0.2cm}
    \caption{Visual comparisons between FastPhysGS and other physics-based 3DGS-MPM simulation and video generation methods. 
    Our method achieves better dynamic visual  performance while exhibiting the lowest simulation memory and the fastest execution speed.}
    \vspace{-0.3cm}
    \label{fig:compare1}
\end{figure*}

\section{Experiment}
\label{sec:experment}

\subsection{Implementation Details}
\textbf{Experimental Setup.} 
Our method is deployed in the PyTorch \cite{Paszke2019PyTorchAI} and Taichi \cite{taichi} environments. 
We compare our method against four
physics-based 3DGS simulation methods using the same dataset and hardware: 
PhysGaussian \cite{xie2024physgaussian}, DreamPhysics\cite{huang2025dreamphysics},
Physics3D \cite{liu2024physics3d} and 
OmniPhysGS \cite{linomniphysgs}. 
Additionally, we also compare with SOTA video generation model CogVideoX-5B \cite{yangcogvideox}, and physics-aware large video model Veo-3.1 \cite{deepmind2025veo}.
All experiments are conducted on a single NVIDIA RTX A6000 GPU with 48 GB memory, 
with each method simulating and rendering 150 frames.

\textbf{Datasets.} 
We evaluate on three public datasets from \cite{xie2024physgaussian, liu2024physics3d, zhang2024physdreamer} with 3DGS format. 
We additionally synthesize the multi-objects 3DGS dataset (e.g., a bear and a can, a duck and a stone). 
For more details, please refer to the Appendix. 

\subsection{Comparison with Baseline Methods}
We conduct qualitative and quantitative comparisons against other physics-based 3DGS simulation approaches and SOTA video generation models. 
We compare the rendering results under identical 
camera trajectories and force priors. 
The motion patterns include translation, rotation, free fall, collision, tearing, and wind swaying, 
with material properties spanning rubber, sand, liquid, metal, and elastomers.

\textbf{Quantitative Comparison.} 
For quantitative evaluation, we adopt the CLIP Score (CS) \cite{pmlr} to measure the semantic consistency between the frames and prompts, 
and the Aesthetic Score (AS) \cite{laion} to assess the visual fidelity. 
We further follow VideoPhy2 \cite{bansal2025videophy}, and adopt Semantic Adherence (SA) to measure the video-text alignment fidelity, while Physical Commensense (PC) measures whether videos obey the physics laws of the real-world. 
We also compare the capability of automatic parameter optimization, running memory consumption, and simulation time. 
As shown in Table \ref{tab:com}, 
compared to the CogVideoX, the physics-aware Veo achieves higher visual quality with AS 4.90 and physical realism with PC 0.49, yet it struggles to simulate dynamics that align with user intention. 
Regarding dynamic 3DGS simulation methods, 
although PhysGaussian achieves low simulation costs, it requires manual parameter tuning, and yields unsatisfactory text-image alignment and rendering quality with CS 0.267 and AS 4.06. 
In contrast, existing video distillation-based methods (Physics3D, DreamPhysics and OmniPhysGS) achieve higher scores in physical realism evaluation with SA 0.81 and PC 0.55, but incur high memory and computational cost. 

By comparison, our FastPhysGS achieves the best performance with the lowest computational cost, requiring only \textcolor{red}{7} GB memory and \textcolor{blue}{1} minute simulation time. 
These demonstrate potential values for interactive performance. 

\textbf{Qualitative Comparison.} 
As shown in Figure \ref{fig:compare1},  
given the same prompt and 3DGS, along with identical motion patterns and force priors, we compare the rendering results under the same camera pose. 
We present results for single-object, multi-objects, 
and scene-level scenarios, respectively. 
CogVideoX struggles to produce physically plausible dynamics, 
while Veo generates high-fidelity results, it still differs from user intention. 
For example, we aim to generate a vertical tearing motion of the bread, CogVideoX and Veo fail to capture this  interaction, resulting in inconsistent behaviors. 
Among physics-based 3DGS methods, 
OmniPhysGS lacks effective interior filling, leading to unrealistic deformation during collision (e.g., the bear hits a metal can but the can exhibits unnatural collapse). 
Other manual-designed or distillation-based methods face challenges in achieving accurate instance-level simulation. 
For example, when simulating the rotation of aircraft blades, the static fuselage erroneously deforms. 

In comparison, our method benefits from IPF, 
providing a instance-level geometric foundation. 
BGDO allows for correcting initial parameters from VLM with better stability.

\subsection{Ablation Study}\label{sec_aba}
We conduct ablation experiments to evaluate the effectiveness of our proposed modules: 

\begin{figure*}[!t]
    \centering
    \includegraphics[width=0.95\linewidth]{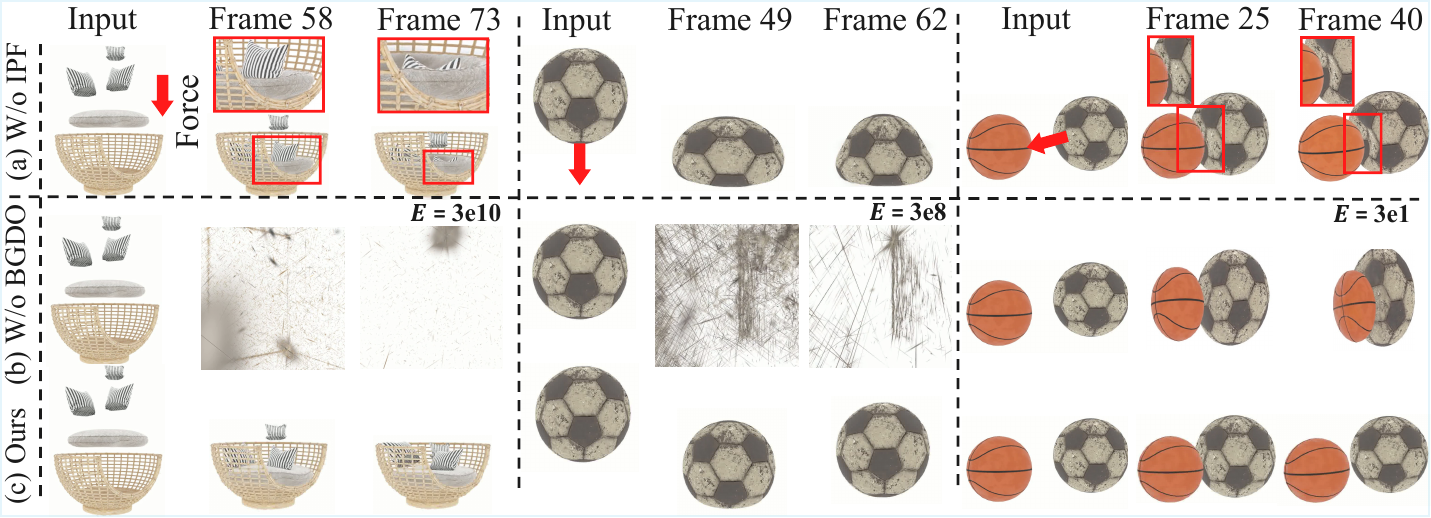}
    \vspace{-0.2cm}
    \caption{Visual comparisons of our ablation studies of FastPhysGS.}
    \vspace{-0.4cm}
    \label{fig:aba}
\end{figure*}

\textbf{W/o IPF.}
As shown in Figure \ref{fig:aba}(a) and Table \ref{tab:aba}, missing particle filling causes internal collapse of 3DGS, 
leading to visually implausible results.
By contrast, our IPF achieves interior filling, providing a geometrically complete 3D structure for stable and realistic MPM simulation.

\begin{figure}[!t]
    \centering
    \includegraphics[width=0.9\linewidth]{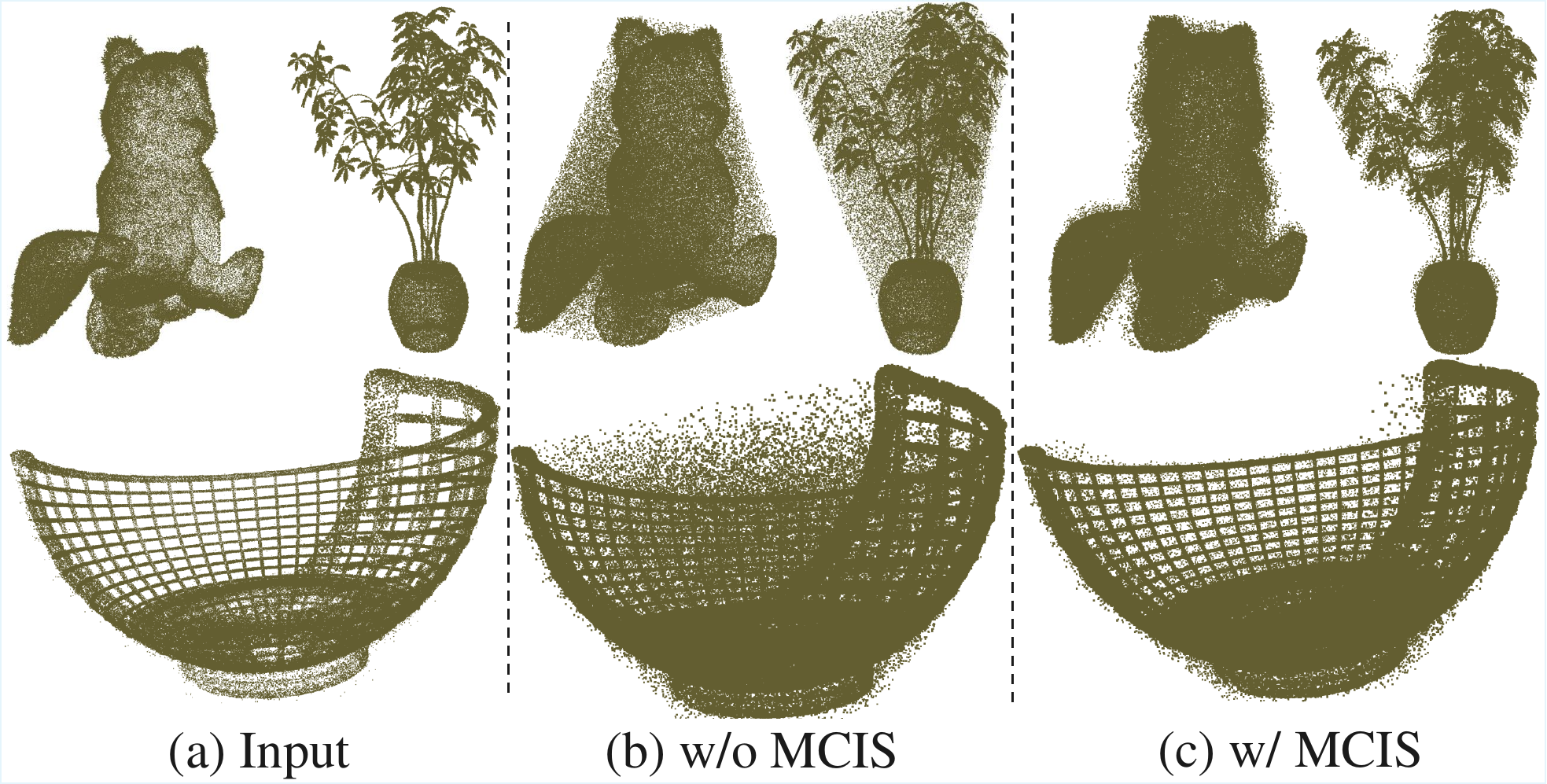}
    \vspace{-0.2cm}
    \caption{Visual ablation study of MCIS.} 
    \label{fig:MCIS}
    \vspace{-0.2cm}
\end{figure}

\textbf{W/o MCIS.}
Figure \ref{fig:MCIS}(b) and Table \ref{tab:aba} show that without MCIS, complex geometries like concave surfaces and irregular curved bodies are challenging to handle, resulting in inaccurate filling. 
MCIS mitigates this issue by modeling importance sampling, achieving correct completion. 

\begin{figure}[!b]
    \centering
    \includegraphics[width=0.82\linewidth]{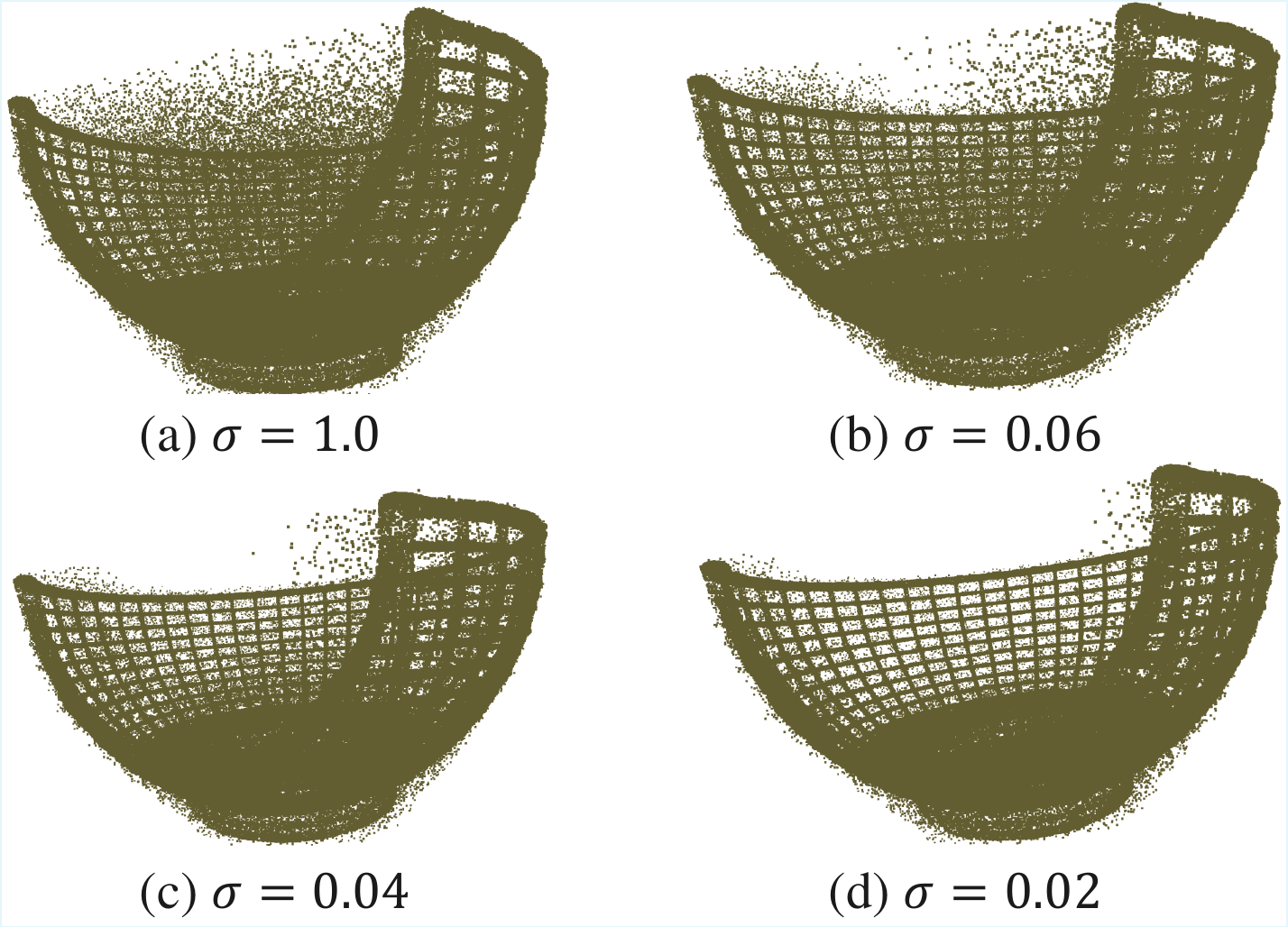}
    \vspace{-0.2cm}
    \caption{Ablation study of standard deviations $\sigma$ of MCIS.}
    \label{fig:std}
\end{figure}

\textbf{$\sigma$ of MCIS.} 
We analyze the influence of different gaussian standard deviations $\sigma$ within MCIS. 
Specifically, we observe the results for $\sigma=$1 (has no effect), 0.06, 0.04, and 0.02, as shown in Figure \ref{fig:std}. 
Smaller $\sigma$ assigns higher weights to points close to cluster centers and suppresses those farther away. 
As a result, sampling becomes highly focused on geometrically salient regions, such as boundaries and non-concave regions. 
In contrast, larger $\sigma$ values flatten the weight distribution, making sampling nearly uniform and diminishing the role of importance. 

\begin{figure}[!t]
    \centering
    \includegraphics[width=0.67\linewidth]{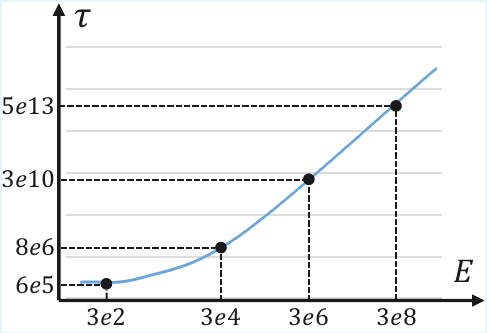}
    \vspace{-0.3cm}
    \caption{Variation of the particle stress $\tau$ with different initial $E$.}
    \label{fig:BGDO}
    \vspace{-0.55cm}
\end{figure}

\textbf{W/o BGDO.}
To assess the robustness of BGDO under corrupted initial  parameters, we set the initial Young’s modulus to 
$3 \times 10^{10}$, $3 \times 10^8$, and $3 \times 10^1$. 
As shown in Figure~\ref{fig:aba}(b) and Table \ref{tab:aba}, these settings lead to varying degrees of simulation failure.  
In contrast, our BGDO stabilizes the system within only two iterations, recovering a reasonable Young's modulus and thereby enabling robust 3DGS simulation.

\begin{table}[htbp]
\centering
\small
\caption{Quantitative metrics of ablation study. 
The full model achieves the best performance, with negligible overhead in running memory (GB) and time (min).}
\vspace{-0.2cm}
\label{tab:aba}
\begin{tabular}{c|c|c|c|c|c|c}
\hline
Method        & CS$\uparrow$  & AS$\uparrow$ & SA$\uparrow$ & PC$\uparrow$  &  Mem. &  Time \\
\hline
w/o IPF      & 0.263  & 4.26   & 0.60  & 0.44 &  \textbf{5}       & 0.8 \\
\hline
w/o MCIS      & 0.285  & 4.31    & 0.72 & 0.52 &  7      & 1  \\
\hline
w/o BGDO       & 0.217  & 3.34   & 0.57  & 0.36 &  7       &  \textbf{0.5}  \\
\hline
Ours          & \textbf{0.292}  & \textbf{4.71}  & \textbf{0.87} & \textbf{0.61}    & 7        & 1  \\
\hline
\end{tabular}
\vspace{-0.5cm}
\end{table}

\textbf{Optimization stability of BGDO.} 
To further validate the stability of BGDO, we evaluate the stress $\tau$ under different initial parameters. 
As illustrated in Figure~\ref{fig:BGDO}, we change only the $E$ while keeping other experiment conditions identical, and plot over 60 sets of results. 
It shows that when $E$ is small, the gradient effect weakens and deformation dominates, while a large $E$ causes a sharp gradient increase. 
This phenomenon prevents the optimization from diverging into false directions, further enhancing the robustness of BGDO. 
For more ablation studies of DBSCAN clustering radius $r$ and material compatibility, 
please refer to the appendix.

\section{Conclusion}
\label{sec:conclusion}
We propose FastPhysGS for efficient and robust 
physics-based 3DGS simulation. 
IPF addresses instability caused by hollow 3DGS structures, while handling complex geometry filling and diverse object instances. 
BGDO achieves rapid material parameters optimization. 
Experiments show the superior performance of FastPhysGS, 
enabling practical real-time applications and interactive dynamic simulation. 



\bibliography{example_paper}
\bibliographystyle{icml2026}

\newpage
\appendix
\onecolumn


\section{Material Point Method}
\label{sec:mpm}

\subsection{MPM Algorithm} 
The Material Point Method (MPM) simulates the materials behavior by discretizing the continuum into particles and updating their properties in a hybrid Eulerian-Lagrangian way. 
MPM naturally handles large deformation and complex interactions with high robustness. 
The algorithm is summarized as follows: 

\textbf{Particle to Grid Transfer.}
In this step, mass $m$ and momentum $m\mathbf{v}$ are transferred from particles to grid nodes by distributing the particle properties to nearby points. 
The accumulated mass and momentum at grid node $ i $ at time step $ n $ are computed as:
\begin{align}
m_i^n &= \sum_p w_{ip}^n m_p, \\
m_i^n \mathbf{v}_i^n &= \sum_p w_{ip}^n m_p \left( \mathbf{v}_p^n + \mathbf{C}_p^n (\mathbf{x}_i - \mathbf{x}_p^n) \right),
\end{align}
where $ w_{ip}^n $ denotes the weighting function between particle $ p $ and grid point $ i $, $ m_p $ is the particle mass, $ \mathbf{v}_p^n $ is the particle velocity, $ \mathbf{x}_p^n $ is the particle position, and $ \mathbf{C}_p^n $ represents the local gradient correction term.

\textbf{Grid Update.}
The grid velocities are updated based on external force and interaction with neighboring particles:
\begin{equation}
\mathbf{v}_i^{n+1} = \mathbf{v}_i^n - \frac{\Delta t}{m_i} \sum_p \tau_p^n \nabla w_{ip}^n V_p^0 + \Delta t \mathbf{f},
\end{equation}
where $ \Delta t $ is the time step, $ \tau_p^n $ is the stress tensor at particle $ p $, $ V_p^0 $ is the reference volume, $ \mathbf{f} $ is the external force, and $ m_i $ is the total mass accumulated at grid node $ i $.

\textbf{Grid to Particle Transfer.}
Velocities are interpolated back from the grid to the particles, updating their states:

\begin{align}
\mathbf{v}_p^{n+1} &= \sum_i \mathbf{v}_i^{n+1} w_{ip}^n, \\
\mathbf{x}_p^{n+1} &= \mathbf{x}_p^n + \Delta t \mathbf{v}_p^{n+1}, \\
\mathbf{C}_p^{n+1} &= \frac{4}{\Delta x^2 } \sum_i w_{ip}^n \mathbf{v}_i^{n+1} (\mathbf{x}_i^n - \mathbf{x}_p^n)^T, \\
\nabla \mathbf{v}_p^{n+1} &= \sum_i \mathbf{v}_i^{n+1} \nabla w_{ip}^n{}^T, \\
\tau_p^{n+1} &= \boldsymbol{\tau}(\mathbf{F}_{\mathbf{E}}^{n+1}, \mathbf{F}_{\mathbf{N}}^{n+1}),
\end{align}
where $ \Delta x $ is the Eulerian grid spacing, $ \mathbf{F}_{\mathbf{E}} $ and $ \mathbf{F}_{\mathbf{N}} $ represent the elastic and normal deformation gradients, respectively, and $ \boldsymbol{\tau} $ is the constitutive stress model.

\subsection{Energy Constitutive Models}
We integrate expert-designed constitutive models, which describe several representative materials including 
elasticity (e.g., rubber, branches, and cloth), 
plasticity (e.g., snow, metal, and clay), 
viscoelasticity (e.g., honey and mud), 
and fluidity (e.g., water, oil, and lava). 

\subsection*{Fixed Corotated Elasticity}

Following \citet{stomakhin2012energetically}, we define fixed corotated elasticity as:
\begin{equation}
\mathbf{P}(\mathbf{F}) = 2\mu (\mathbf{F} - \mathbf{R}) + \lambda J (J - 1) \mathbf{F}^{-T},
\label{eq:fixed_corotated}
\end{equation}
where $\mathbf{R}$ is the rotation matrix obtained from the polar decomposition $\mathbf{F} = \mathbf{R}\mathbf{S}$, $J = \det(\mathbf{F})$ is the volume change, and $\mu$, $\lambda$ are the Lamé parameters. This model is well-suited for simulating rubber-like materials due to its ability to capture large rotations with minimal shear distortion.

\subsection*{Neo-Hookean Elasticity}

Based on \cite{bonet1997nonlinear}, the Neo-Hookean elasticity is defined as:
\begin{equation}
\mathbf{P}(\mathbf{F}) = \mu \left( \mathbf{F} - \mathbf{F}^{-T} \right) + \lambda \log(J) \, \mathbf{F}^{-T},
\label{eq:neo_hookean}
\end{equation}
this formulation assumes isotropic hyperelastic behavior and is particularly effective in modeling spring-like elastic responses, such as those seen in soft tissues or idealized linear elastic solids.

\subsection*{StVK Elasticity}

We adopt the St. Venant-Kirchhoff (StVK) model \cite{barbivc2005real}, which is given by:
\begin{equation}
\mathbf{P}(\mathbf{F}) = \mathbf{U} \left( 2\mu \boldsymbol{\Sigma}^{-1} \ln \boldsymbol{\Sigma} + \lambda \operatorname{tr}(\ln \boldsymbol{\Sigma}) \, \boldsymbol{\Sigma}^{-1} \right) \mathbf{V}^T,
\label{eq:stvk}
\end{equation}
where $\mathbf{F} = \mathbf{U} \boldsymbol{\Sigma} \mathbf{V}^T$ is the singular value decomposition (SVD) of the deformation gradient. The matrices $\mathbf{U}$, $\boldsymbol{\Sigma}$, and $\mathbf{V}$ represent the left stretch, singular values, and right rotation, respectively. The StVK model captures both elastic and plastic behaviors and is suitable for simulating materials such as sand and metals, especially when combined with plasticity models.

\subsection*{Identity Plasticity}

The identity plasticity model assumes no plastic deformation, meaning the material behaves purely elastically. This is commonly used for idealized elastic materials: 
\begin{equation}
\psi(\mathbf{F}) = \mathbf{F}, 
\label{eq:identity_plasticity}
\end{equation}
this model serves as a baseline for comparison with more complex plasticity formulations.

\subsection*{Drucker-Prager Plasticity}

Following \cite{drucker1952soil, klar2016drucker}, we define Drucker-Prager plasticity using a return mapping based on the singular value decomposition (SVD) of the deformation gradient. Let $\mathbf{F} = \mathbf{U} \boldsymbol{\Sigma} \mathbf{V}^T$, where $\boldsymbol{\Sigma}$ contains the singular values. Define $\bm{\epsilon} = \log(\boldsymbol{\Sigma})$, which represents the logarithmic strain. The return mapping is then given by:
\begin{align}
& \psi(\mathbf{F})  = \mathbf{U} \mathcal{Z}(\boldsymbol{\Sigma}) \mathbf{V}^T,
\\
& \mathcal{Z}(\boldsymbol{\Sigma})  =
\begin{cases}
\mathbf{I}, & \text{if } \sum(\bm{\epsilon}) > 0, \\
\boldsymbol{\Sigma}, & \text{if } \delta\gamma \leq 0 \text{ and } \sum(\bm{\epsilon}) \leq 0, \\
\exp\left( \bm{\epsilon} - \delta\gamma \frac{\hat{\bm{\epsilon}}}{\|\hat{\bm{\epsilon}}\|} \right), & \text{otherwise},
\end{cases}
\label{eq:drucker_prager}
\end{align}
where $\delta\gamma$ is the plastic update parameter, $\hat{\bm{\epsilon}}$ is the deviatoric part of $\bm{\epsilon}$, and $\sum(\bm{\epsilon})$ denotes the sum of the diagonal entries of $\bm{\epsilon}$. This model is particularly suitable for simulating granular materials such as snow and sand, due to its ability to capture dilatancy and yield under pressure.

\subsection*{von Mises Plasticity}

Based on \cite{hu2018moving}, von Mises plasticity is defined through a similar SVD-based return mapping:
\begin{align}
& \psi(\mathbf{F}) = \mathbf{U} \mathcal{Z}(\boldsymbol{\Sigma}) \mathbf{V}^T,
\\
& \mathcal{Z}(\boldsymbol{\Sigma}) =
\begin{cases}
\boldsymbol{\Sigma}, & \text{if } \delta\gamma \leq 0, \\
\exp\left( \bm{\epsilon} - \delta\gamma \frac{\hat{\bm{\epsilon}}}{\|\hat{\bm{\epsilon}}\|} \right), & \text{otherwise}.
\end{cases}
\label{eq:von_mises}
\end{align}
this formulation captures isotropic yielding and is widely used to simulate ductile materials such as metals and clay, where plastic flow occurs under shear stress.

\subsection*{Fluid Plasticity}

We adopt the fluid plasticity model from \citet{stomakhin2014augmented}, which treats the material as incompressible and fluid-like. The return mapping is defined as:

\begin{equation}
\psi(\mathbf{F}) = J^{1/3} \mathbf{I},
\label{eq:fluid_plasticity}
\end{equation}
where $J = \det(\mathbf{F})$ is the volume ratio. This model effectively removes all elastic deformation and enforces a uniform deformation state, making it suitable for simulating highly fluid-like materials such as water and lava.

\section{More Method Details of FastPhysGS}

\subsection{Elasticity Models and Stress Sensitivity to $E$ }
\label{app:elasticity}

All elasticity models in our framework compute the first Piola-Kirchhoff stress $ \boldsymbol{\tau} = \partial \Phi / \partial \mathbf{F}$ from the deformation gradient $\mathbf{F} \in \mathbb{R}^{3\times3} $, where $ \Phi $ is the strain energy density. 

Crucially, each model expresses $ \boldsymbol{\tau}$ as a linear function of the Lam\'e parameters: 
\begin{equation}
\mu = \frac{E}{2(1+\nu)}, \quad \lambda = \frac{E\nu}{(1+\nu)(1-2\nu)},
\end{equation}
which are themselves linear in Young’s modulus $E$. 
This linearity implies that the stress scales approximately linearly with  $E $, i.e., $ \boldsymbol{\tau} \propto E$, 
making $ \|\boldsymbol{\tau}\| $ a natural proxy for material stiffness.

We implement five models: 
\begin{itemize}
    \item \textbf{SigmaElasticity (Hencky/Logarithmic Strain)}: Uses principal logarithmic strains $ \boldsymbol{\varepsilon} = \log(\boldsymbol{\Sigma}) $ , where $\mathbf{F} = \mathbf{U}\boldsymbol{\Sigma}\mathbf{V}^\top $ . 
    The stress is:   
\begin{equation}
    \boldsymbol{\tau} = \mathbf{U} \operatorname{diag}(2\mu \boldsymbol{\varepsilon} + \lambda \operatorname{tr}(\boldsymbol{\varepsilon})\mathbf{1}) \mathbf{U}^\top.
\end{equation}
    This model is energetically consistent and well-suited for large deformations.

    \item \textbf{CorotatedElasticity}: Removes rigid rotation via polar decomposition  $ \mathbf{R} = \mathbf{U}\mathbf{V}^\top $ :
\begin{equation}
    \boldsymbol{\tau} = 2\mu (\mathbf{F} - \mathbf{R})\mathbf{F}^\top + \lambda J(J - 1)\mathbf{I},
\end{equation}
    where  $ J = \det(\mathbf{F}) $ . The deviatoric term captures shear, while the volumetric term enforces near-incompressibility.

    \item \textbf{StVKElasticity}: Based on Green--Lagrange strain  $ \mathbf{E} = \frac{1}{2}(\mathbf{F}^\top\mathbf{F} - \mathbf{I}) $ :
\begin{equation}
    \boldsymbol{\tau} = 2\mu \mathbf{F}\mathbf{E} + \lambda J(J - 1)\mathbf{I}.
\end{equation}
    Accurate for moderate strains but may stiffen unrealistically under large compression.

    \item \textbf{FluidElasticity}: Sets  $ \mu = 0 $ , yielding purely volumetric response:
\begin{equation}
    \boldsymbol{\tau} = \lambda J(J - 1)\mathbf{I}.
\end{equation}
    Suitable for liquids or highly dissipative materials.

    \item \textbf{VolumeElasticity}: Uses Mie--Gr\"uneisen equation of state:
\begin{equation}
        \boldsymbol{\tau} = \kappa \left( J - J^{-\gamma+1} \right) \mathbf{I}, \quad \kappa = \tfrac{2}{3}\mu + \lambda,\ \gamma=2.
\end{equation}
        Better stability under extreme compression.
\end{itemize}

In all cases,  $ \partial \boldsymbol{\tau} / \partial E $  exists and is non-zero, enabling gradient-based optimization of  $ E $ . However, numerical instabilities—especially near  $ J \approx 0 $  or during plastic yielding—cause  $ \|\nabla_E \|\boldsymbol{\tau}\|\| $  to span many orders of magnitude ( $ 10^8 $ – $ 10^{20} $ ), necessitating robust normalization.

\subsection{Plasticity Models and Their Impact on Optimization}
\label{app:plasticity}

Plasticity modifies  $ \mathbf{F} $  after the elastic stress computation, thereby indirectly affecting the stress-deformation relationship used in BGDO. We support four models: 
\begin{itemize}
    \item \textbf{IdentityPlasticity}: No modification; purely elastic.

    \item \textbf{SigmaPlasticity}: Enforces unilateral incompressibility by clamping  $ J = \det(\mathbf{F}) \in [0.05, 1.2] $  and resetting  $ \mathbf{F} = J^{1/3} \mathbf{I} $. This suppresses unrealistic expansion and stabilizes simulation, but reduces sensitivity to  $ E $  in highly compressed regions.

    \item \textbf{VonMisesPlasticity}: Yields when deviatoric strain exceeds yield stress  $ \sigma_y $ :
\begin{equation}
    \text{if } \|\boldsymbol{\varepsilon}_{\text{dev}}\| > \frac{\sigma_y}{2\mu}, \text{ then project } \boldsymbol{\varepsilon} \text{ onto yield surface}.
\end{equation}
    Since  $ \mu \propto E $ , higher  $ E $  raises the yield threshold, making the material appear more elastic—this nonlinearity is captured by BGDO’s stress gradient signal.

    \item \textbf{DruckerPragerPlasticity}: Pressure-dependent yielding:
\begin{equation}
    f(\boldsymbol{\tau}) = \|\boldsymbol{\tau}_{\text{dev}}\| + \alpha \operatorname{tr}(\boldsymbol{\tau}) - c \leq 0,
\end{equation}
    where  $ \alpha $  depends on friction angle  $ \phi $ , and  $ c $  is cohesion. Compression increases resistance to yielding. Because  $ \operatorname{tr}(\boldsymbol{\tau}) \propto E $ , the yield condition itself depends on  $ E $ , creating a complex but differentiable coupling that BGDO exploits.
\end{itemize}

Critically, \textbf{all plasticity corrections are applied after stress computation}, so the gradient  $ \nabla_E \|\boldsymbol{\tau}\| $  remains well-defined and reflects pre-plastic elastic behavior—exactly the signal needed for stiffness calibration.

\subsection{Deformation Guidance via Frobenius Norm}
\label{app:frobenius}

The Frobenius norm of the post-step deformation gradient deviation,
\begin{equation}
\delta = \|\mathbf{F}' - \mathbf{I}\|_F = \sqrt{\sum_{i=1}^3 \sum_{j=1}^3 \left(F'_{ij} - \delta_{ij}\right)^2},
\end{equation}
quantifies the magnitude of local strain accumulated over a single MPM substep, where  $ \mathbf{F}' $  is the updated deformation gradient after one explicit integration step and  $ \delta_{ij} $  is the Kronecker delta. Although this measure is not fully rotation-invariant (unlike invariants of  $ \mathbf{C} = \mathbf{F}^\top \mathbf{F} $ ), the small time step size  $ \Delta t $  in our simulation ensures that rotational components in  $ \mathbf{F}' $  are close to identity, i.e.,  $ \mathbf{F}' \approx \mathbf{I} + \nabla \mathbf{u} $  with displacement gradient  $ \nabla \mathbf{u} $  small. Consequently,  $ \|\mathbf{F}' - \mathbf{I}\|_F $  approximates the Euclidean norm of the infinitesimal strain tensor and serves as a practical proxy for visible deformation.

We set a target value  $ \delta_{\text{target}} = 0.1 $ , chosen empirically to balance visual responsiveness and physical plausibility:
\begin{itemize}
    \item If  $ \delta \ll \delta_{\text{target}} $ , the material exhibits negligible motion—indicating excessive stiffness—so we reduce the penalty on softness (effectively encouraging a decrease in  $ E $ );
    \item If  $ \delta \gg \delta_{\text{target}} $ , the material may sag, collapse, or exhibit numerical instability—suggesting insufficient stiffness—so we increase  $ E $ .
\end{itemize}
This deformation-based signal resolves an inherent ambiguity in stress-only optimization: two materials with different  $ E $  can produce similar stress magnitudes under static loading but exhibit drastically different dynamic responses. By incorporating  $ \delta $ , BGDO aligns parameter calibration with perceptual motion cues.

\subsection{Gradient Compression and Log-Space Update}
\label{app:grad_compress}

The raw gradient of the stress norm with respect to the log-scale modulus,
\begin{equation}
\mathbf{g}_\tau = \nabla_{\log E} \|\boldsymbol{\tau}\| = \frac{\partial \|\boldsymbol{\tau}\|}{\partial \log E},
\end{equation}
exhibits extreme dynamic range due to the nonlinear dependence of $ \boldsymbol{\tau} $ on $ \mathbf{F} $  and the exponential mapping $ E = \exp(\log E) $. In regions near element inversion ( $ J = \det(\mathbf{F}) \to 0 $ ) or plastic yield, $ \|\boldsymbol{\tau}\|$ can change by orders of magnitude with tiny changes in $ E $, leading to $ \|\mathbf{g}_\tau\| \in [10^8, 10^{20}] $ in practice. Direct use of such gradients causes optimizer divergence or dominance by outlier particles.

To stabilize training, we apply a smooth, monotonically increasing compression function:
\begin{equation}
\tilde{g}_\tau = \log(1 + |\mathbf{g}_\tau|).
\end{equation}
This transformation:
\begin{itemize}
    \item Compresses the dynamic range; 
    \item Preserves gradient sign and monotonicity—larger stress sensitivity still yields larger update magnitude;
    \item Reduces the influence of extreme outliers during spatial averaging across particles or temporal averaging across key frames.
\end{itemize}

The final update operates in log-space:
\begin{equation}
\log E \gets \log E -  \mathcal{G}, \quad \text{where } \mathcal{G} = \frac{1}{|\mathcal{K}|} \sum_{k \in \mathcal{K}} \left[ \eta \cdot \tilde{g}_{\tau}^{(k)} - (1 - \eta) \cdot \delta^{(k)} \right],
\end{equation}
with key frame set $\mathcal{K}$, and balancing weight $ \eta \in [0,1]$. Updating  $\log E $ guarantees  $E = \exp(\log E) > 0$  and induces multiplicative updates on $E$ , 
which are more natural for scale parameters than additive ones. This formulation enables robust convergence within 1–2 iterations while using only three stored simulation snapshots, achieving minimal memory overhead.

\subsection{Algorithm and Limitation}
The BGDO optimization flow, summarized in Algorithm \ref{alg:bgdo}, decouples the causal forward simulation from the non-causal parameter update. 
By restricting gradient computation to three key frames and leveraging compressed stress gradients with Frobenius-norm deformation guidance, 
it achieves stable and memory-efficient material calibration. 

\begin{algorithm}[htbp]
\caption{Bidirectional Graph Decoupling Optimization (BGDO)}
\label{alg:bgdo}
\begin{algorithmic}[1]
\Require 
    Initial Young’s modulus $\log E$, material labels $\{l_k\}$, 
    total frames $T$, key frames $\mathcal{K} = \{0, t, n\}$,
    target deformation magnitude $\delta_{\text{target}}$, balancing weight $w \in [0,1]$
\Ensure Optimized  $ \log E $ 

\Statex \textbf{Stage 1: Gradient-Free Forward Simulation}
\State Initialize particle states  $ (\mathbf{x}, \mathbf{v}, \mathbf{C}, \mathbf{F}) $ 
\State  $ \texttt{frame\_buffer} \gets \emptyset $ 
\For{ $ \text{frame} = 0 $  to  $ T - 1 $ }
    \If{ $ \text{frame} \in \mathcal{K} $ }
        \State Store snapshot:  $ \texttt{frame\_buffer} \gets \texttt{frame\_buffer} \cup \{(\mathbf{x}, \mathbf{v}, \mathbf{C}, \mathbf{F})\} $ 
    \EndIf
    \For{ $ \text{substep} = 1 $  to  $ S $ } \Comment{ $ S $ : MPM substeps per frame}
        \State  $ \boldsymbol{\tau} \gets \textsc{Elasticity}(\mathbf{F},\, \log E.\text{detach}()) $ 
        \State  $ (\mathbf{x}, \mathbf{v}, \mathbf{C}, \mathbf{F}) \gets \textsc{MPMStep}(\mathbf{x}, \mathbf{v}, \mathbf{C}, \mathbf{F}, \boldsymbol{\tau}) $ 
        \State  $ \mathbf{F} \gets \textsc{Plasticity}(\mathbf{F},\, \log E.\text{detach}()) $ 
    \EndFor
\EndFor

\Statex \textbf{Stage 2: Gradient-Based Backward Optimization}
\If{ $ \texttt{frame\_buffer} \neq \emptyset $ }
    \State  $ \mathbf{g}_{\text{total}} \gets \mathbf{0} $ 
    \For{each  $ (\mathbf{x}_i, \mathbf{v}_i, \mathbf{C}_i, \mathbf{F}_i) \in \texttt{frame\_buffer} $ }
        \State  $ \mathbf{F}_i \gets \mathbf{F}_i.\text{requires\_grad}() $ 
        \State  $ \boldsymbol{\tau}_i \gets \textsc{Elasticity}(\mathbf{F}_i,\, \log E) $ 
        \State  $ (\_, \_, \_, \mathbf{F}_i') \gets \textsc{MPMStep}(\mathbf{x}_i, \mathbf{v}_i, \mathbf{C}_i, \mathbf{F}_i, \boldsymbol{\tau}_i) $ 
        \State  $ \mathbf{F}_i' \gets \textsc{Plasticity}(\mathbf{F}_i',\, \log E) $ 
        
        \Statex \textit{// Stress gradient signal (prevent over-stiffness)}
        \State  $ s_i \gets \|\boldsymbol{\tau}_i\|_F $ 
        \State  $ \mathbf{g}_{\tau} \gets \nabla_{\log E}\, s_i $ 
        \State  $ \tilde{g}_{\tau} \gets \log(1 + |\mathbf{g}_{\tau}|) $ 
        
        \Statex \textit{// Deformation guidance (prevent over-softness)}
        \State  $ \delta_i \gets \|\mathbf{F}_i' - \mathbf{I}\|_F $ 
        \State  $ \tilde{g}_{\delta} \gets \delta_i - \delta_{\text{target}} $ 
        
        \State  $ \mathbf{g}_{\text{total}} \gets \mathbf{g}_{\text{total}} + w \cdot \tilde{g}_{\tau} - (1 - w) \cdot \tilde{g}_{\delta} $ 
    \EndFor
    \State  $ \log E \gets \log E - \frac{1}{|\mathcal{K}|} \mathbf{g}_{\text{total}} $ 
\EndIf

\State \Return  $ \log E $ 
\end{algorithmic}
\end{algorithm}

\begin{table*}[htbp]
\centering
\caption{List of Symbols and Their Meanings}
\label{tab:symbols_only}
\begin{tabular}{lll}
\toprule
\textbf{Symbol} & \textbf{Type} & \textbf{Meaning} \\
\midrule
 $ \bm{\mu} $  & Vector  & Center of a 3D Gaussian \\
 $\bm{\Sigma}$  & Matrix & Covariance matrix of a 3D Gaussian \\ 
 $ \bm{\Sigma'} $  &  Matrix & Projected 2D covariance in screen space \\ 
 $ \alpha_i $  & Scalar   & Opacity of the  $ i $ -th Gaussian \\
 $ N $  & Integer & Number of Gaussians \\
\midrule
 $ m $  & Scalar & Particle mass in MPM \\
 $ \textbf{f} $  & Vector & Force vector in MPM \\
 $ \Delta(\cdot, t) $  & Function & Time-dependent spatial deformation map \\
 $ \bm{F}_i(t) $  &  Matrix & Deformation gradient of particle  $ i $  at time  $ t $  \\
 $ t $  & Scalar & Simulation time step \\
\midrule
 $ \bm{\mathcal{P}} $  & Set & Original set of 3DGS particle positions \\
 $ \bm{p}_i $  & Vector & Position of  $ i $ -th original Gaussian ( $ = \bm{\mu}_i $ ) \\
 $ K $  & Integer & Number of object instances after clustering \\
 $ \bm{\mathcal{C}_k} $  & Set & Point cluster for instance  $ k $  \\
 $ l_k $  & Integer & Instance identifier for cluster  $ k $  \\
 $ r $  & Scalar & DBSCAN neighborhood radius \\
 $ \bm{c}_j^{(k)} $  & Vector &  $ j $ -th point in cluster  $ \bm{\mathcal{C}_k} $  \\
 $ \bm{\mathcal{H}_k} $  & Mesh & Convex hull of cluster  $ k $  \\
 $ \bm{\mathcal{B}_k} $  & AABB & Axis-aligned bounding box of  $ \bm{\mathcal{H}_k} $  \\
 $ \bm{b}_{\min}^{(k)}, \bm{b}_{\max}^{(k)} $  & Vectors & Min/max corners of AABB for instance  $ k $  \\
 $ \bm{\mathcal{Q}_k} $  & Set & Candidate filling points \\ 
 $ M $  & Integer & Number of candidate points \\
 $ \text{occ}(\bm{q}; \bm{\mathcal{H}_k}) $  & Scalar & Occupancy probability \\
 $ \bm{\mathcal{Q}_k^{\text{inside}}} $  & Set & Interior candidates  \\
 $ d_j^{(k)} $  & Scalar & Minimal Euclidean distance  \\
 $ \sigma $  & Scalar & Gaussian kernel scale  \\
 $ \epsilon $  & Scalar & Small constant ( $ 10^{-6} $ ) for numerical stability \\
 $ w_j^{(k)} $  & Scalar & Importance weight  \\
 $ p_j^{(k)} $  & Scalar & Normalized sampling probability  \\
 $ n_k $  & Integer & Number of filled points selected for instance  $ k $  \\
 $ \bm{\mathcal{S}_k} $  & Set & Final sampled interior points for instance  $ k $  \\
 $ \bm{\mathcal{P}_{\text{filled}}} $  & Set & All filled points \\
 $ \bm{\mathcal{P}_{\text{all}}} $  & Set & Full particle set \\
\midrule
 $ \rho $  & Scalar & Material density  \\
 $ \nu $  & Scalar & Poisson’s ratio  \\
 $ \Phi $  & Function & Energy density model  \\
 $ E $  & Scalar & Young’s modulus  \\
 $ \bm{I} $  & Image & Rendered frames \\
 $ \bm{\mathcal{X}} $  & Set & 3D positions of all Gaussians \\
 $ \bm{v} $  & Vector & Particle velocity in MPM \\
 $ \bm{C} $  & Matrix & Affine velocity field \\
 $ \boldsymbol{\tau} $  & Tensor & First Piola–Kirchhoff stress \\
 $ \bm{f_0}, \bm{f_t}, \bm{f_n} $  & Frames & Key frames (initial, mid, final) \\
 $ g_{\tau} $  & Scalar & Average stress gradient \\
 $ d_{\bm{F}} $  & Scalar & Average Frobenius norm of $\bm{F} - \mathbf{I}$  \\
 $ g $  & Scalar & Log-compressed stress gradient \\
 $ \mathcal{G} $  & Scalar & Combined optimization signal \\
 $ \eta $  & Scalar & Adaptive weight \\
\bottomrule
\end{tabular}
\end{table*}

\textbf{Limitation}
For highly fine-grained complex scenes, 
interior filling may be limited by segmentation accuracy, thereby affecting material precision. 
Adopting a large segmentation model could alleviate this issue. 
In the future, we will focus on more general representations and physical interactions.

\section{More Experimental Results}
\label{sec:exp}

\subsection{More Setup Details}
To synthesize the multi-objects dataset, we collect public single-object datasets and then modify the underlying 3DGS code to enable the fusion of single-object scenes.
In MPM, spatial normalization is applied to a $1\times1\times1$ volume, and the object center is placed at (0.5, 0.5, 0.5). 
The simulation runs for 150 frames, the prompt is generated by a large language model, and all other parameters, such as camera poses and viewing angles are determined by the input 3DGS scene. 

\subsection{Ablation Study of DBSCAN}
\begin{figure}[htbp]
    \centering
    \includegraphics[width=0.85\linewidth]{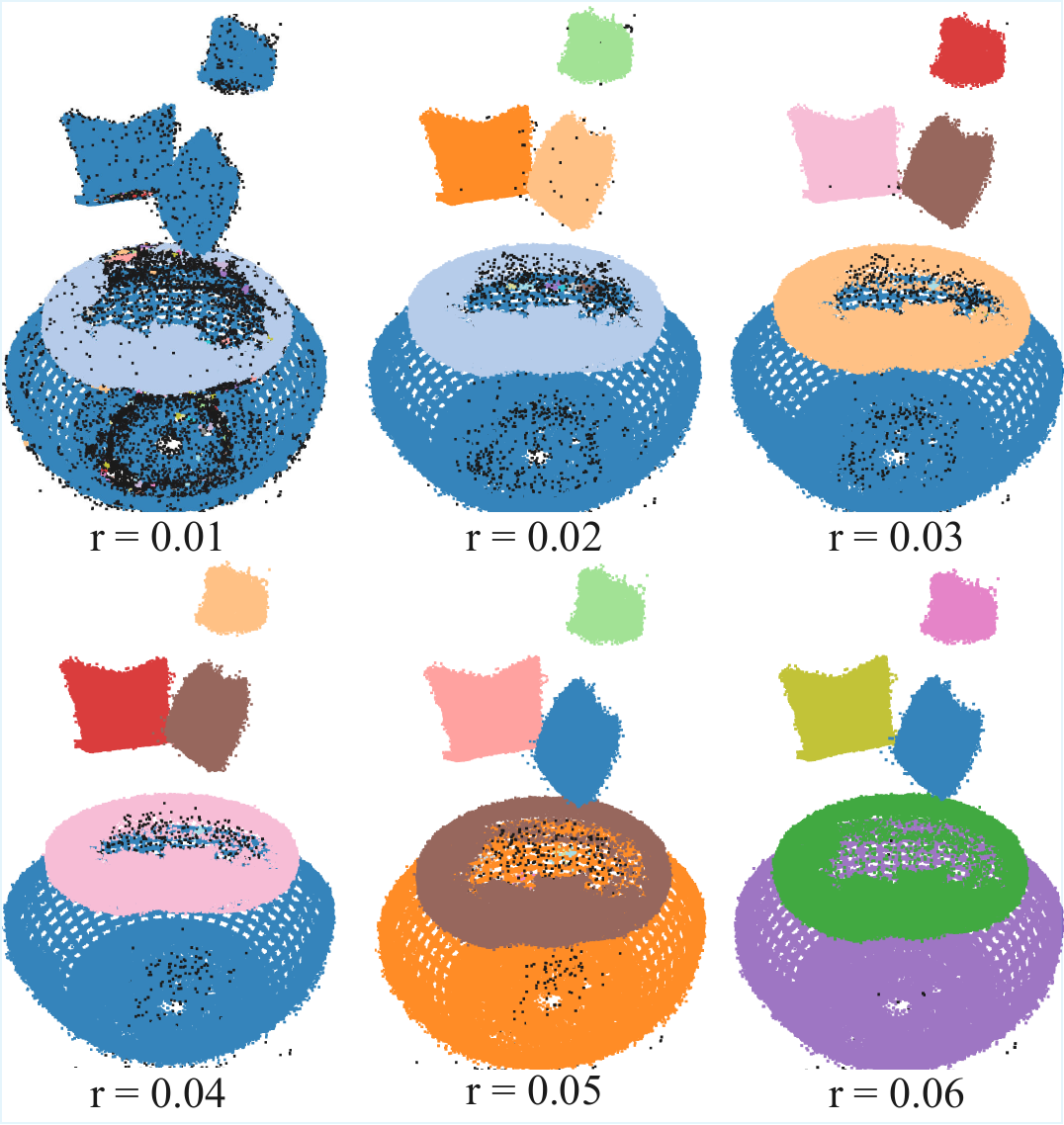}
    \caption{Ablation study of different r of DBSCAN.}
    \label{fig:dbscan}
\end{figure}

To more intuitively illustrate the impact of different radius values on DBSCAN clustering, we visualize the segmentation results obtained with radii ranging from 0.01 to 0.06 in Figure \ref{fig:dbscan}. 
As can be observed, a radius of 0.01 exhibits almost no discriminative capability, yielding very poor segmentation. 
As the radius increases, the segmentation quality gradually improves. 
Starting from a radius of 0.05, the segmentation becomes accurate, albeit at the cost of increased computational time. 
In practice, we typically select a radius in the range of 0.05 to 0.07.

\subsection{More Visualization Results}

\begin{figure*}[htbp]
    \centering
    \includegraphics[width=0.97\linewidth]{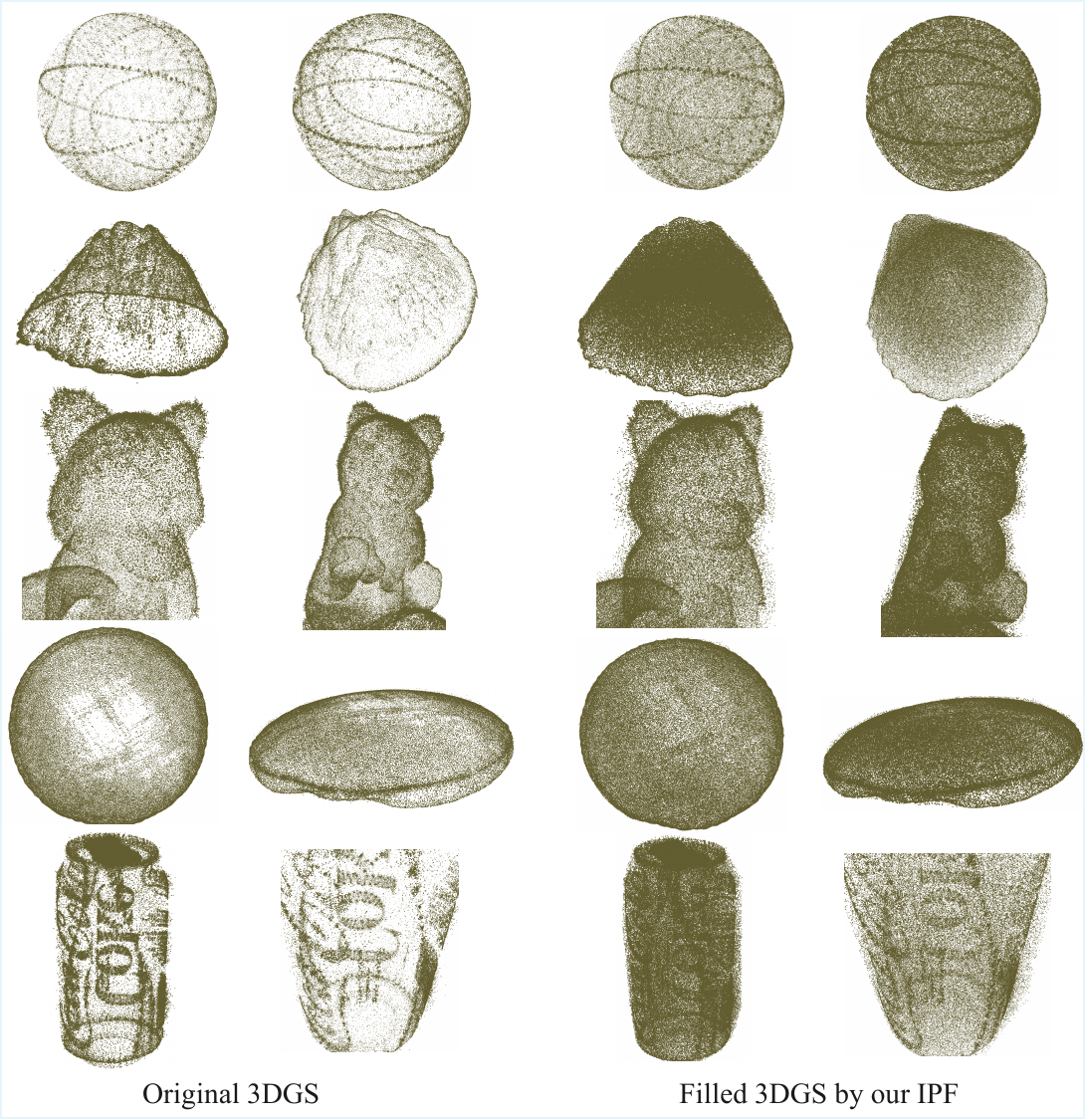}
    \caption{More visualized results of our IPF,  demonstrating its ability to achieve stable filling across diverse structures.}
    \label{fig:fillres}
\end{figure*}

\begin{figure*}[htbp]
    \centering
    \includegraphics[width=0.99\linewidth]{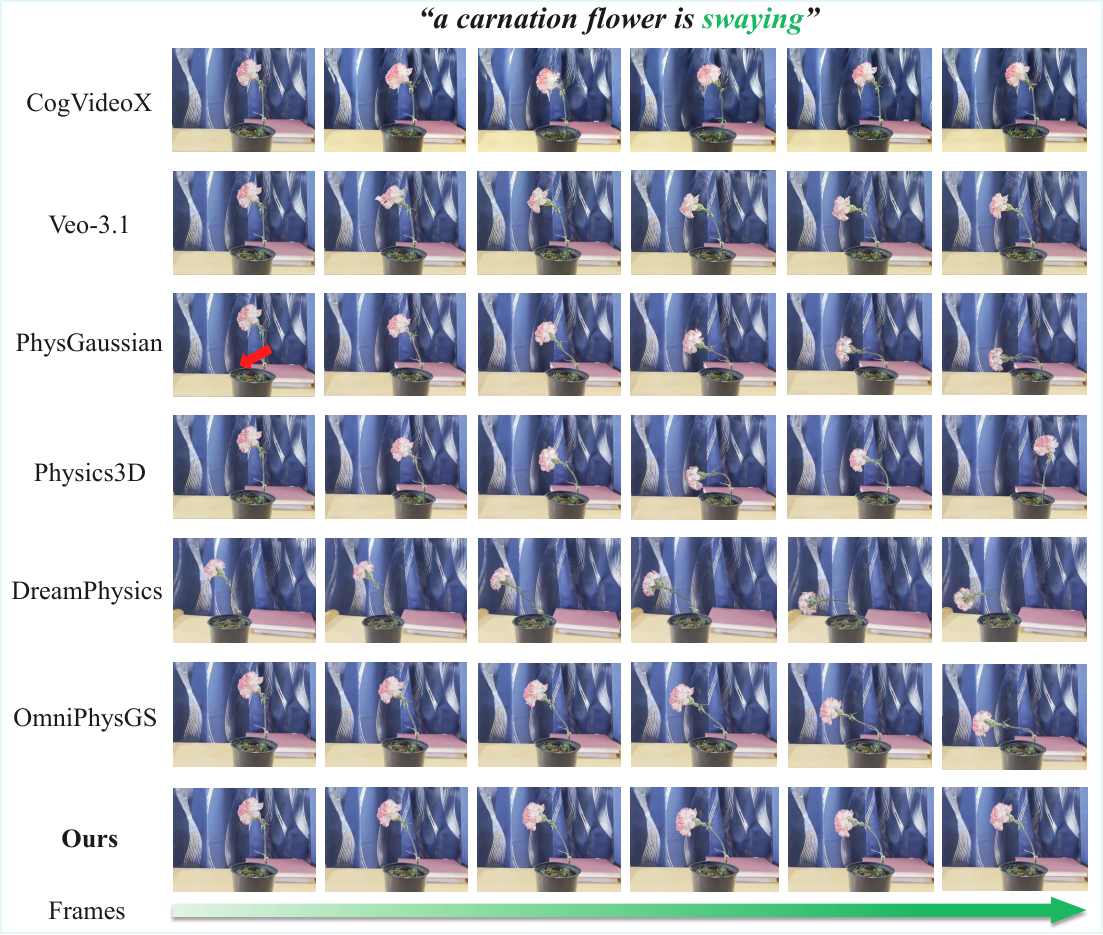}
    \caption{More visual comparisons of  FastPhysGS and other baseline methods. Given the same prompt and force prior, our method produces more realistic dynamics. For example, this flower swings to a certain extent and then swings back.} 
    \label{fig:res1}
\end{figure*}

\begin{figure*}[htbp]
    \centering
    \includegraphics[width=0.98\linewidth]{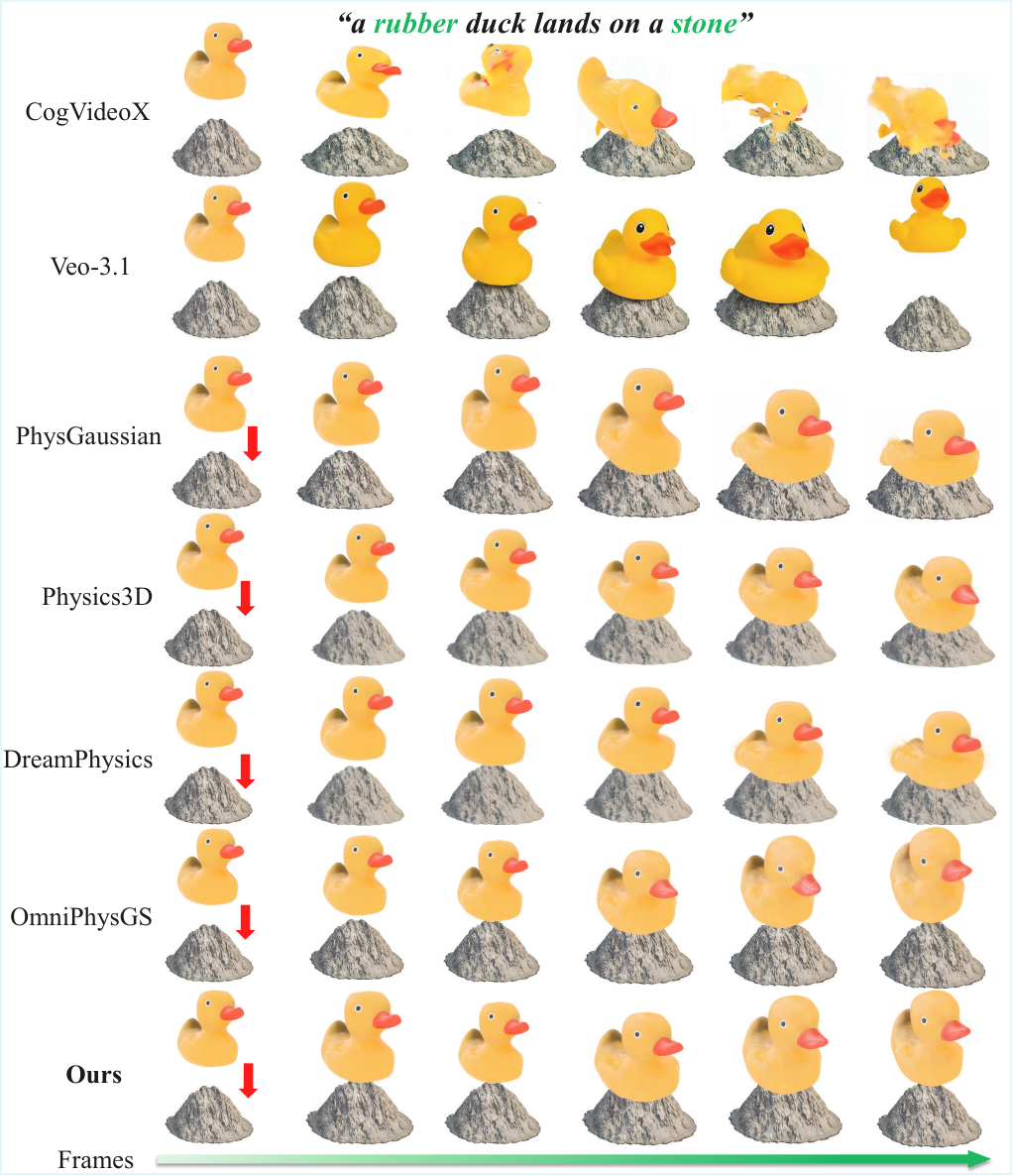}
    \caption{More visual comparisons of  FastPhysGS and other baseline methods. Given the same text description and force priors, our method generates more realistic dynamic effects. 
    For example, the rubber duck lands on a stone and bounces back.} 
    \label{fig:res2}
\end{figure*}

\begin{figure*}[htbp]
    \centering
    \includegraphics[width=0.99\linewidth]{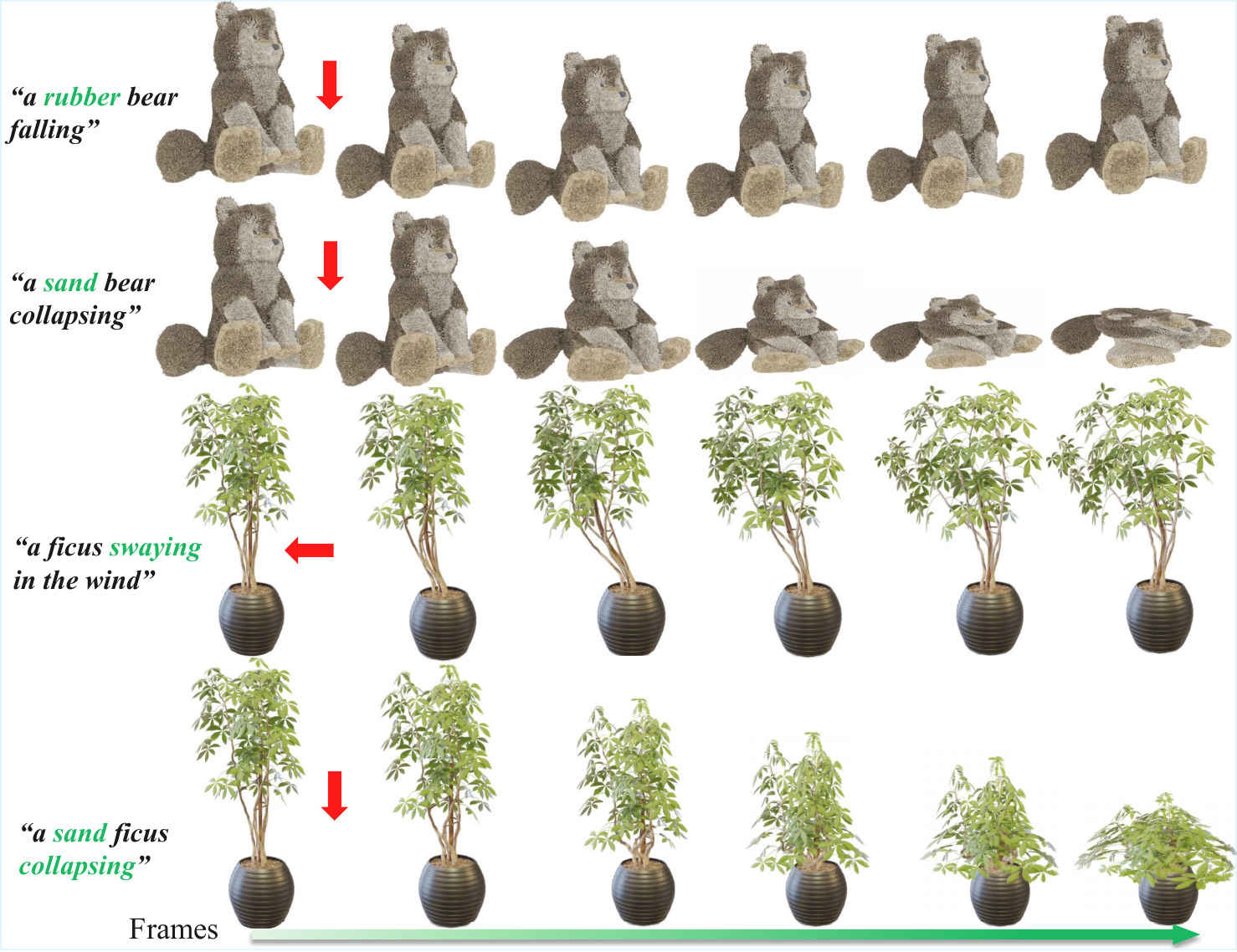}
    \vspace{-0.2cm}
    \caption{We further demonstrate the capability to simulate different material properties of FastPhysGS. 
    For instance, in the bear scene, our method produces bouncy behavior for a rubber material, as well as collapse effects for sand or liquid materials. 
    Meanwhile, ficus achieves soft-body swinging and collapse.}
    \vspace{-0.3cm}
    \label{fig:res3}
\end{figure*}

\begin{figure*}[htbp]
    \centering
    \includegraphics[width=0.99\linewidth]{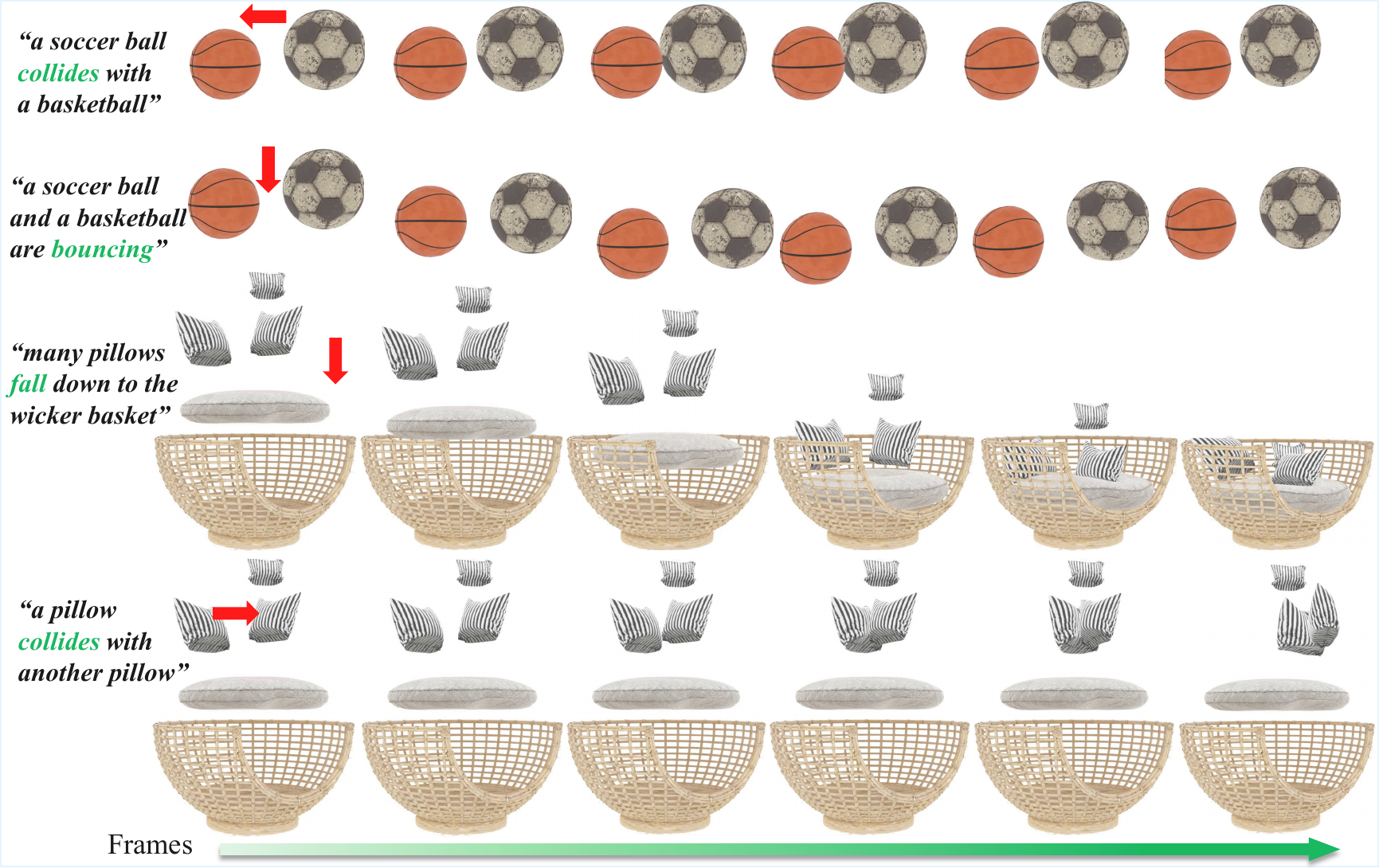}
    \caption{We further demonstrate the ability to simulate diverse motion priors. 
    For example, it can handle the free fall and multi-collisions of objects, such as balls and pillows.}
    \label{fig:res4}
\end{figure*}

\begin{figure*}[htbp]
    \centering
    \includegraphics[width=0.99\linewidth]{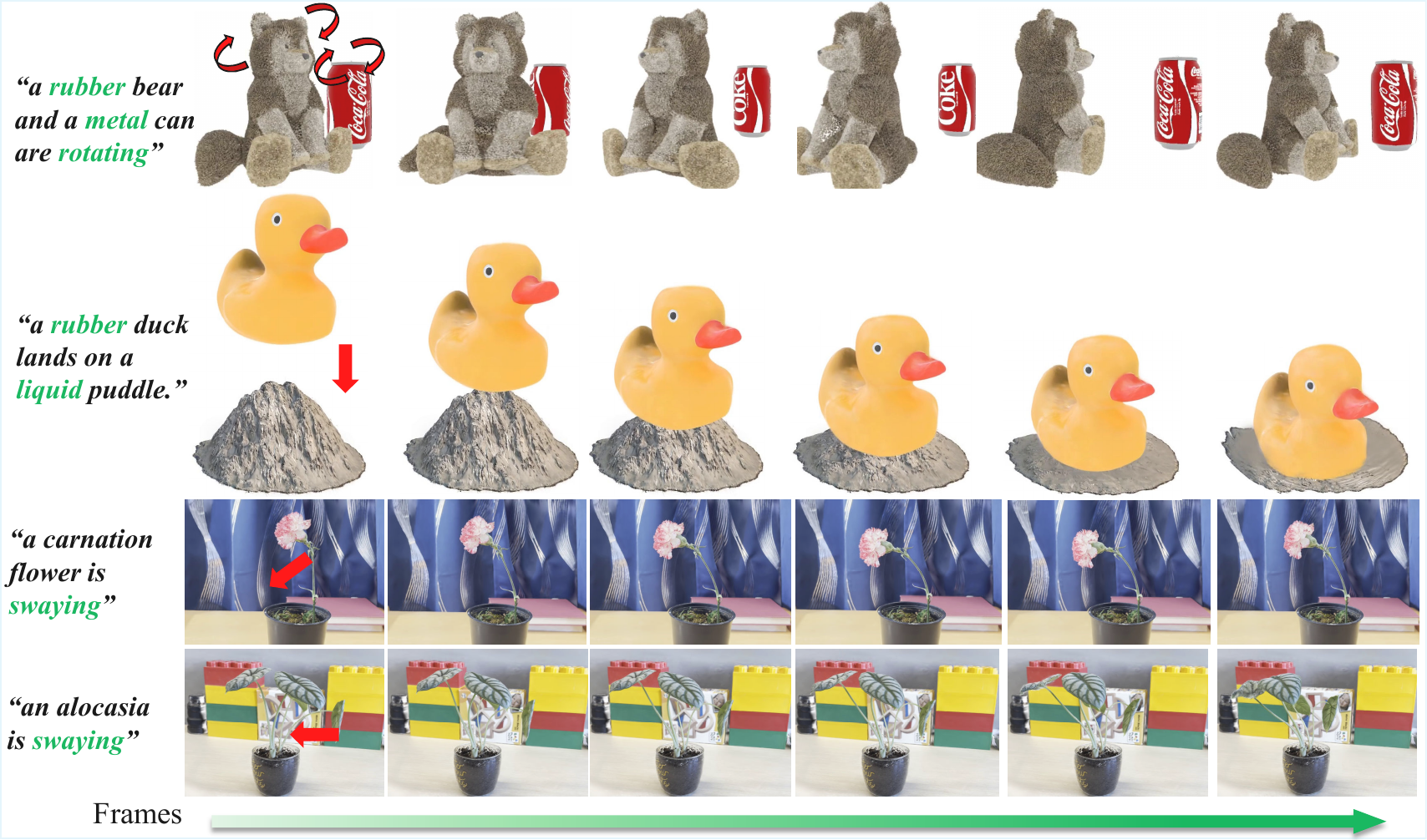}
    \caption{We further demonstrate the ability to simulate diverse motion priors and material properties of different scenes.}
    \label{fig:res5}
\end{figure*}

We present more visualization results in Figures \ref{fig:fillres} to \ref{fig:res5}, showcasing a rich variety of material simulations, such as rubber, elastomers, sand, liquids, flexible materials, soft bodies, and rigid bodies. 
These further demonstrate the superiority and practicality of our FastPhysGS.


\end{document}